\documentclass[10pt]{article} 

\usepackage[accepted]{rlj} 

\usepackage{amssymb}            
\usepackage{mathtools}          
\usepackage{mathrsfs}           
\usepackage{graphicx}           
\usepackage{subcaption}         
\usepackage[space]{grffile}     
\usepackage{url}                
\usepackage{lipsum}             

\usepackage{float}
\usepackage[utf8]{inputenc} 
\usepackage[T1]{fontenc}    
\usepackage{hyperref}       
\usepackage{url}            
\usepackage{booktabs}       
\usepackage{amsfonts}       
\usepackage{nicefrac}       
\usepackage{microtype}      
\usepackage{xcolor}         
\usepackage{microtype}
\usepackage{graphicx}
\usepackage{booktabs} 
\usepackage{array}
\usepackage{multirow}
\usepackage{caption}  
\usepackage{enumitem}
\usepackage{makecell}
\usepackage{wrapfig}
\usepackage{natbib}

\usepackage{color}
\usepackage{xcolor}
\newcommand{\minisection}[1]{\vspace{0.05in} \noindent {\bf #1} \ }

\DeclareMathOperator{\EX}{\mathbb{E}}

\title{Human-Like Goalkeeping in a Realistic Football Simulation: a Sample-Efficient Reinforcement Learning Approach}

\setrunningtitle{Human-Like Goalkeeping in a Realistic Football Simulation: a Sample-Efficient RL Approach}

\author{Alessandro Sestini\textsuperscript{1}, Joakim Bergdahl\textsuperscript{1}, Jean-Philippe Barrette-LaPierre\textsuperscript{1}, Florian Fuchs\textsuperscript{1}, Brady Chen\textsuperscript{2}, Fabio Zinno\textsuperscript{2}, Michael Jones\textsuperscript{2}, Linus Gisslén\textsuperscript{1}}

\emails{\{asestini, jbergdahl, jbarrettelapierre, ffuchs, bradychen, fzinno, michaelj, lgisslen\}@ea.com}

\affiliations{
$^{1}$\textbf{Electronic Arts (EA), Stockholm, Sweden}\\
$^{2}$\textbf{EA Sports, Vancouver, Canada}
}

\contribution{
    We introduce a sample-efficient off-policy DRL paradigm for a sample-constrained continuous control problem, such as the development of a video game. The approach builds on prior work on sample efficiency and combines known techniques to improve learning speed and stability, making machine learning more viable for video game production.
    }
    {
    The constantly evolving nature of games makes it difficult for developers to deploy DRL solutions that require weeks to produce results~\citep{autoplayers}. Prior work on sample-efficient DRL~\citep{srspr, rlpd, bbf} has provided ways to improve learning speed -- in terms of sample efficiency and wall-clock time. Our results demonstrate that combining these techniques yields a viable approach for applying RL in a game production setting. Through ablation, we isolate the effect of each component on sample efficiency, stability, and final performance, highlighting the most impactful improvements in recent work.
    }

\contribution{
    We propose an expert-guided fine-tuning procedure based on symmetric replay sampling, enabling targeted behavioral correction while preserving previously acquired skills. 
    }
    {
    We re-purpose replay across experiments~\citep {acrossexp}, a symmetric sampling scheme that leverages offline data from previous experiments. While the original paper used a majority of online data, we show that this approach can be applied with a minority of online data, thus enabling accessible fine-tuning.
    }

\contribution{
    We test the algorithm in the game \textit{EA SPORTS FC 25} to train and improve the goalkeeper's positioning AI. Our agent outperforms the game's built-in AI by 10\% in ball saving rate. We show that our method trains agents 50\% faster compared to standard DRL methods such as SAC~\citep{sac}. Qualitative evaluation by domain experts demonstrates that our approach produces more human-like gameplay than hand-crafted agents. 
    }
    {
    As a testament to the approach's impact, the method has been adopted for the most recent release of \textit{EA SPORTS FC}. Our findings and demonstration results for \textit{EA SPORTS FC 25} open the door for other game studios to introduce RL-based solutions even with limited compute resources.
    }


\summary{
While several high-profile video games have served as testbeds for Deep Reinforcement Learning (DRL), the high computational demands of standard DRL approaches -- driven by their sample inefficiency -- pose a significant barrier to adoption for game studios seeking to develop authentic AI behaviors. This paper proposes a sample-efficient DRL method tailored for training and fine-tuning agents in industrial settings such as the video game industry. Our method improves sample efficiency of value-based DRL by leveraging pre-collected data and increasing network plasticity. We evaluate our method through training and shipping a goalkeeper agent in \textit{EA SPORTS FC}, one of the best-selling football simulations today. 
}

\begin{document}

\maketitle  

\begin{abstract}
While several high-profile video games have served as testbeds for Deep Reinforcement Learning (DRL), this technique has rarely been employed by the game industry for crafting authentic AI behaviors. Previous research focuses on training super-human agents with large models, which is impractical for game studios with limited resources aiming for human-like agents. This paper proposes a sample-efficient DRL method tailored for training and fine-tuning agents in industrial settings such as the video game industry. Our method improves sample efficiency of value-based DRL by leveraging pre-collected data and increasing network plasticity. We evaluate our method training a goalkeeper agent in \textit{EA SPORTS FC 25}, one of the best-selling football simulations today. Our agent outperforms the game's built-in AI by 10\% in ball saving rate. Ablation studies show that our method trains agents 50\% faster compared to standard DRL methods. Finally, qualitative evaluation from domain experts indicates that our approach creates more human-like gameplay compared to hand-crafted agents. As a testament to the impact of the approach, the method has been adopted for use in the most recent release of the series.
\end{abstract}
\section{Introduction}
\label{sec:introduction}

\begin{figure*}
\centering
\includegraphics[width=0.95\linewidth]{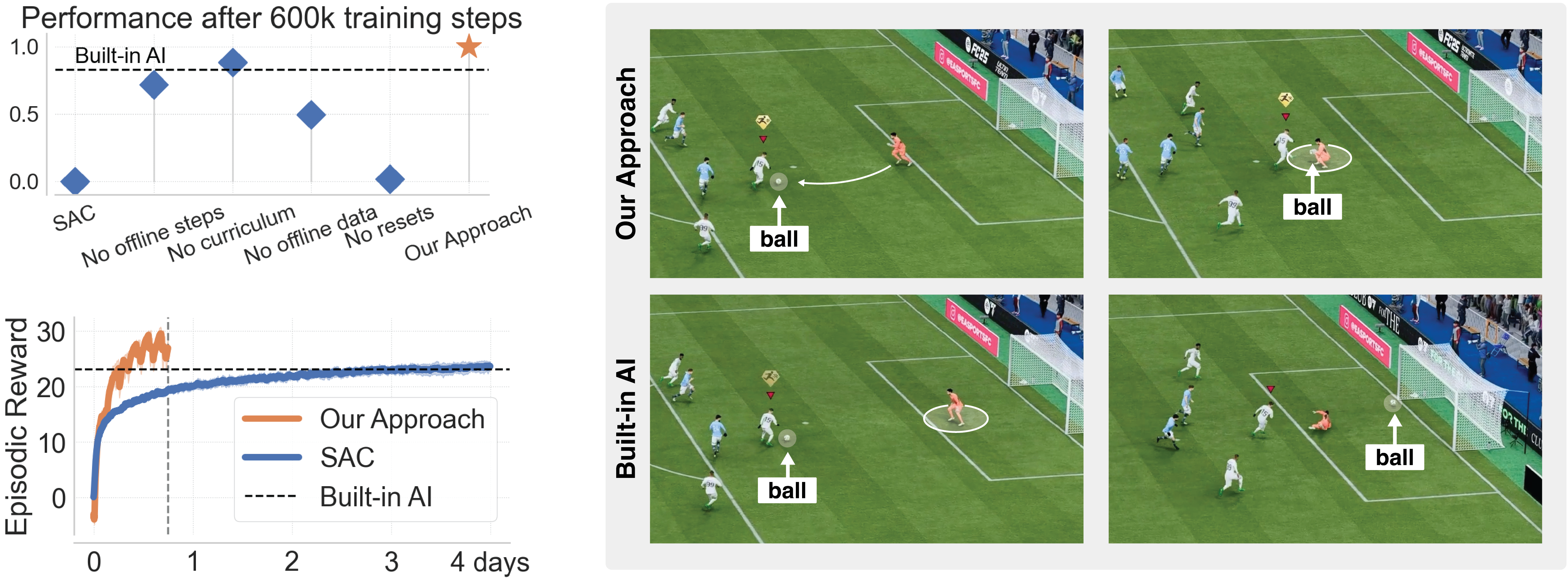}
\caption{\textbf{Main results of our approach}. \textbf{Top-left}:  our approach compared to removing the variations we add on top of the standard SAC algorithm. \textbf{Bottom-left}: training time compared to standard SAC algorithm. Our method outperforms built-in AI in less than one day. The standard SAC algorithm is able to match the performance of the built-in AI, but only after 4 days of training. \textbf{Right}: an example showcasing the behavioral differences between our agent (top) and the built-in AI (bottom) in the same situation. Our agent is more proactive and better understands the current situation, anticipating the shot. Quoting a professional goalkeeper: \textit{``the goalkeeper plays it really well! The keeper looks for opportunities to steal ground as the striker enters the box.''}}
\label{fig:results_summary}
\end{figure*}

Deep Reinforcement Learning (DRL) research has demonstrated significant potential in areas such as robotics~\citep{schulmanrobotic,berkeleyrobotic}, control of nuclear fusion plasma in a tokamak~\citep{tokamak}, and design of faster sorting algorithms~\citep{deepmindsorting}. At the same time, the video game industry experienced significant technological progress -- in areas such as computer graphics, physics, and design -- that led to the development of more complex and immersive game experiences. Despite this progress, designing Artificial Intelligence (AI) systems to manage Non-Player Characters (NPCs) remains a complex element of the creative process that significantly influences the quality of games~\citep{unwieldy}. Games provide a natural testbed for DRL research, and the application of DRL in this context has shown great promise. Notable examples such as OpenAI Five~\citep{openaifive} and AlphaStar~\citep{alphastar} show how DRL can outperform professional players in complex video games. These successes have accelerated the deployment of this technique in commercial video games, with examples such as GT Sophy~\citep{sophy}, Naruto Mobile~\citep{shukai}, and Arena Breakout~\citep{arena}, while still not yet widely employed in the industry. 

The success of the aforementioned approaches depends on extensive training, requiring a massive amount of online environment interactions and large neural networks. Although DRL can find well-performing policies with enough interactions, applying it in real-world scenarios is complex. For instance, the constantly evolving nature of games makes it difficult for developers to deploy solutions that require days or weeks to produce results~\citep{autoplayers}. Therefore, such application requires a method that:
(1) outperforms classical hand-crafted AI systems both in terms of numerical performance as well as perceived human-likeness;
(2) trains quickly, both in terms of sample efficiency and wall-clock time; and
(3) is easy to adjust without retraining agents from scratch. \looseness=-1

This paper proposes a sample-efficient method for training human-like AI in video games that is computationally efficient enough to be practically applied in production. Moreover, the method allows for easy modifications to the agent without restarting the training procedure from scratch. We analyze recent research on sample-efficient and offline RL~\citep{bbf, rlpd, efficientzero} and showcase the techniques we combined and extended for developing our method. We evaluate our approach on \emph{EA SPORTS FC 25}, a AAA football simulation game. In particular, we train the goalkeeper's positioning system. This choice stems from the need to improve the hand-crafted, non-realistic, and complex-to-maintain AI. Determining where to position the goalkeeper to better save a goal or anticipate opponents is complex, and manually crafting the decision-making process is challenging. Not to mention, developing a system that mimics real human players is usually a time-consuming and difficult task.

In summary, the key contributions are: (1) we introduce a training paradigm for sample-constrained continuous-control DRL, combining replay-ratio scheduling, hard network resets, offline optimization phases, and curriculum learning to improve learning speed and stability; (2) we demonstrate how suboptimal offline data can be leveraged to bootstrap online learning without limiting final performance; (3) we propose an expert-guided fine-tuning procedure based on symmetric replay sampling that enables targeted behavioral correction while preserving previously acquired skills; (4) we provide extensive ablation studies isolating the effect of each component on sample efficiency, stability, and final performance; and (5) we validate the approach in a large-scale commercial environment (\emph{EA SPORTS FC 25}) under realistic production constraints, and demonstrate improved sample efficiency on standard MuJoCo benchmarks. A summary of the results is shown in Figure~\ref{fig:results_summary}. Although the main focus of this paper is on game development, the findings transfer to other areas where sample efficiency and human-likeness are fundamental challenges, as shown in additional benchmark domains (see Appendix~\ref{app:mujoco}).

\section{Related Work}
\label{sec:related}
The challenge of improving sample efficiency in DRL has been gaining interest from the research community, highlighting its importance for successful applications. Additionally, an increasing number of game studios are attempting to deploy DRL agents in games. 

\minisection{Sample-efficient DRL.} 
DRL requires extensive interaction with the environment to reach the desired performance, which becomes impractical in real-world applications where such interactions are expensive, time-consuming, or risky. \citet{drq} use data augmentation to design a sample-efficient DRL method, while \citet{spr} use a self-supervised temporal consistency loss with data augmentation; EfficientZero is a model-based RL algorithm with self-supervised learning to learn a temporally consistent environment model and use it to correct off-policy value targets~\citep{efficientzero}.  However, most of these approaches focus on learning with limited data, often at the cost of increasing computational resource needs, resulting in time-consuming training. 
Recent results suggest that scaling the replay ratio factor -- the number of optimization steps over environment steps --  is a simple but effective approach for enhancing sample efficiency when combined with periodic network resets~\citep{srspr, bbf}. Moreover, recent literature in offline RL such as RLPD~\citep{rlpd} and SDBG~\citep{sdbg} shows great promise in using offline data for boosting online policies. In this work, we combine and extend several of the latest advances in sample efficiency to address the challenges defined in Section~\ref{sec:introduction}.

\minisection{DRL for video games.}
Recent advancements in DRL have led to impressive results in complex games such as StarCraft II, Dota 2, and Gran Turismo 7~\citep{alphastar, openaifive, sophy}. However, these approaches aim to create super-human agents with extraordinary resources. For instance, the work by \citet{openaifive} took months to train an agent that could beat professional players in the game Dota 2. As stated by \citet{unwieldy}, the video game industry does not need agents built to “beat the game”, but rather to produce human-like behaviors. Similarly, many other fields such as robotics or autonomous driving do not require super-human agents, but rather agents that can better fit the context. In video games, some notable examples include: DeepCrawl, an effective DRL system that is able to create a variety of NPC behaviors for a published roguelike game~\citep{deepcrawl}; the work by \citet{arena} that trained a DRL agent for the game Arena Breakout; and Sh\=ukai, a practical DRL algorithm specifically tailored for commercial fighting games~\citep{shukai}. Although the goal of the cited works is to develop an agent suitable for integration into a real product rather than solely outperforming human performance, none of the papers address the practical implications of training a sample-inefficient DRL agent in a sample-restricted setting.

\section{Methodology}
\label{sec:algorithmology}
In here, we first define preliminaries useful for understanding the subsequent sections. Second, we describe the algorithm we use and our evaluation framework. Finally, we show how we enable human-in-the-loop for easy fine-tuning of an under-performing agent in specific scenarios.


\subsection{Preliminaries}
\label{sec:preliminaries}
An RL setting is commonly formalized as a Markov Decision Process (MDP) which consists of a tuple $\langle S, A, R, P, \gamma \rangle$, where $S$ is the space state, $A$ the action space, $P: S \times A \rightarrow S$ the transition function, $R:S \times A \rightarrow \mathbb{R}$ the reward function and $\gamma \in [0, 1)$ the discount factor. A policy $\pi$ is formalized as a function that maps states to a distribution of actions. The goal is to find an optimal policy that maximizes the sum of expected discounted reward. We represent this objective with an action-value function $Q^\pi(s_t,a_t) = \EX_{\pi, P}[\sum^{N}_{k=1}\gamma^k r_{t + k + 1} \; | \; s_t \in S, a_t \in A]$, where $N$ is the time horizon. In the case of offline RL, the agent does not interact with the environment, but learns from a fixed offline dataset~\citep{offlinesurvey}. Offline RL typically assumes access to $N$ previously-collected transitions $D = \{s_t^i, a_t^i, r_t^i, s_{t+1}^i\}_{i=1}^N$, which are gathered using a policy $\pi_o$. The goal remains to find the optimal policy that maximizes the expected discounted reward. 

In this work, we build our method on top of the value-based Soft Actor-Critic (SAC)~\citep{sac} algorithm. We give a thorough description of SAC in Appendix~\ref{app:sac}. Most modern DRL value function-based approaches store the agent's experience with the environment in a replay buffer, potentially keeping that data for the entire training period. In this regard, the replay ratio -- the ratio between the number of gradient updates and the number of environment steps-- plays a crucial role. For instance, the original DQN algorithm~\citep{dqn} uses a replay ratio of $0.25$, while recent sample-efficient algorithms such as BBF~\citep{bbf} or SR-SPR~\citep{srspr} use 8 and 16. \looseness=-1

\subsection{EA SPORTS FC 25}
\label{sec:environment}
We evaluate our method on the game \emph{EA SPORTS FC 25}, developed by Electronic Arts (EA). The game is part of the \textit{EA SPORTS FC} series, with new entries released every year. \textit{EA SPORTS FC 24} reached a peak of 21 million weekly active users in fiscal year 2024. Figure~\ref{fig:results_summary} shows screenshots of the game. The game is a fully physics-based football simulation, where players play against other humans and in-game AI systems. While the outfield players in the game have advanced, hand-crafted AI, the goalkeeper system suffers from suboptimal behavior. We aim to improve upon this system by leveraging DRL, while respecting the requirements listed in Section~\ref{sec:introduction}. We train our agent using a low-resolution version of the game. This version removes all graphical enhancements unnecessary for training, and it allows for unlocked frame rates, speeding up the simulation by a factor of three. 

\subsection{Algorithm}
\label{sec:algorithm}
Our algorithm extends the base SAC approach by incorporating modifications aimed at improving sample efficiency.
Our approach focuses on training under limited interaction budgets and industrial constraints. Our experiments show that learning efficiency can be improved by dynamically alternating between online interaction, offline optimization, and network reinitialization. Moreover, in Section~\ref{sec:fine_tuning}, we show how combining offline data and online interactions can be used as a fine-tuning approach to improve the performance of the agent and to incorporate expert feedback. To the best of our knowledge, SAC has been the state-of-the-art approach (as well as other off-policy methods) for researchers in video game industry due to its proven sample-efficiency~\cite{sophy,alonso2020deep,bairamian2023minimax}.

Here, we list all the changes applied to the base algorithm.


\begin{figure*}
    \centering 
    \includegraphics[width=0.9\linewidth]{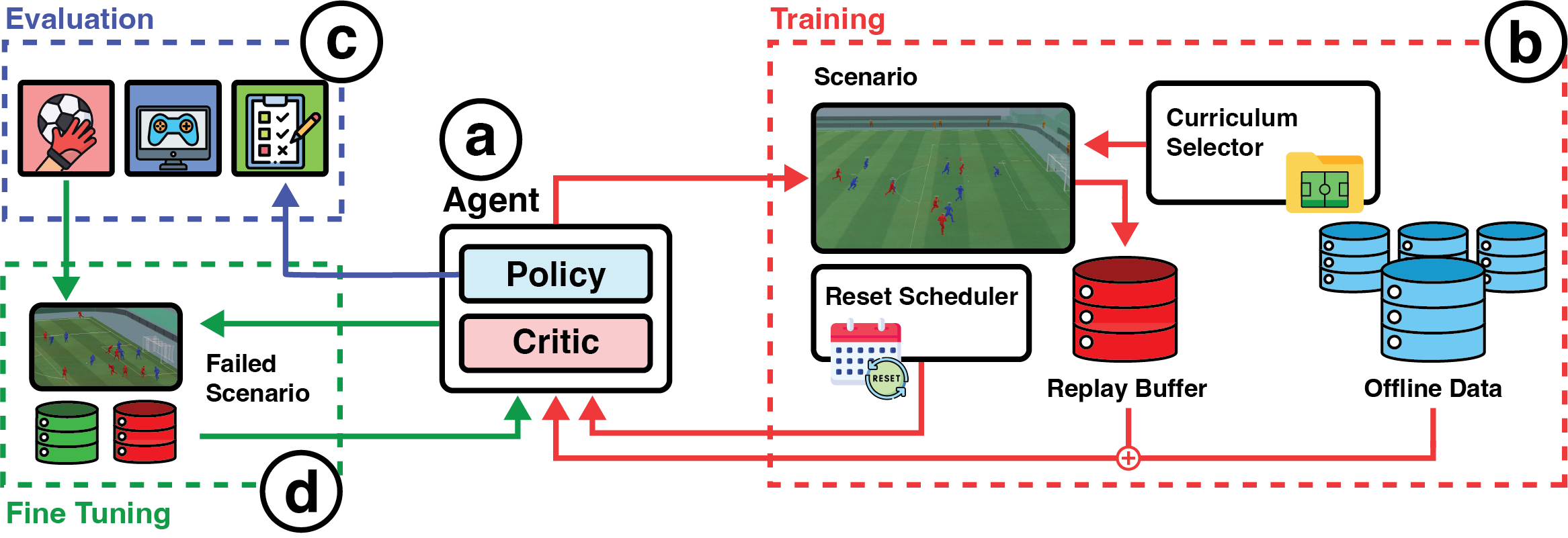}
    \caption{
    \textbf{Overview of the proposed method}. (a) shows the agent composed of the policy and action-value functions as employed by SAC. (b) shows the main training framework, composed of the elements delineated in Section~\ref{sec:algorithm}: \textit{curriculum learning}, \textit{offline data}, and \textit{network resets}. (c) shows the three components of the evaluation framework: \textit{automatic quantitative evaluation}, \textit{human qualitative evaluation}, and \textit{expert-authored test suite}, described in Section~\ref{sec:evaluation}. Finally, (d) shows the fine tuning component. In this, we use only the failed scenario to generate new data but we combine it with the previously collected replay buffer. More details in Section~\ref{sec:fine_tuning}.
    } 
    \label{fig:method}
\end{figure*}

\minisection{Replay ratio and hard reset.} Following the SR-SPR~\citep{srspr} and BBF~\citep{bbf} algorithms, and in contrast to classical approaches~\citep{dqn, sac}, in our experiments we use a replay ratio of 1. In comparison with SR-SPR and BBF, which employ soft resets of the value function, we perform hard resets of both the policy and value function networks every $10^5$ steps. Soft resets involve the reinitialization of a subset of weights, such as the last layers, while hard resets completely reinitialize the network parameters. Moreover, unlike the other two approaches, the moment we apply a hard reset, we increase the replay ratio to $6.4\times10^3$ and train using the current buffer, without allowing other online environment interactions. This is equivalent to performing a vast number of offline updates during the reset. After the offline steps, we resume online training with a replay ratio of 1. In Section~\ref{sec:ablations}, we show how these modifications help improve the sample efficiency of the algorithm while keeping the computational cost low.

\minisection{Scenario-based learning.} A full football match provides only a few salient situations for the goalkeeper. Hence, it is more efficient to train using specific situations. With the help of a domain expert, we define scenarios mimicking real training challenges faced by human goalkeepers. A scenario includes one or more situations. These situations vary in the starting positions of the players; in the particular situation played, e.g. a corner kick; and in the behavior of the other players, e.g. the skill level of the built-in AI for the opponents and teammates. For training the agent using scenarios, we employ curriculum learning~\citep{curriculum}, dividing the scenarios into multiple phases. When we move from one phase to the next, to mitigate the risk of catastrophic forgetting, we reuse a subset of scenarios encountered in previous phases. Example of scenarios and more details on curriculum learning are described in Appendix~\ref{app:curriculum}.

\minisection{Learning with offline data.} The game already features a built-in AI solution for the goalkeeper. We decided to leverage the behavior of the built-in AI to bootstrap the phases of the curriculum. For this goal, we first collect a dataset of $N$ transitions running the built-in AI for each curriculum phase:
$
    D_{\pi_o} = \{s_t^i, a_t^i, r_t^i, s_{t+1}^i\}_{i=1}^N,
$
where $\pi_o$ represents the built-in AI, and $r_t$ is generated from the reward function. Similarly to RLPD~\citep{rlpd}, we then use the symmetric sampling technique, where for each training batch we sample 50\% of the data from the agent's replay buffer $D_\pi$, and the remaining 50\% from $D_{\pi_o}$. We use layer normalization to mitigate catastrophic overestimation. However, unlike RLPD, we do not employ an ensemble of Q-networks. Using multiple Q-networks can improve sample efficiency but requires significant computational resources. In our preliminary experiments, we observed that we could achieve substantial improvements in sample efficiency using only the offline dataset and layer normalization. Furthermore, $D_{\pi_o}$ is sub-optimal. Hence, in order to outperform the built-in AI, we remove $D_{\pi_o}$ after the first reset in each of the curriculum phases.

\subsection{Action and state spaces}
The action space of the agent consists of three continuous values: two values for the relative target position, and one value for the intensity of the movement. We use the same state space for both the agent's policy and action-value functions. These features are directly available from the game engine. We use a total of 110 features, each normalized to the range $[-1, 1]$. The state space is divided into three main components: \textit{goalkeeper features}, a set of $33$ ego-centric values such as relative positions of the ball, relative positions of the goal, and velocities of the goalkeeper; \textit{opponent features}, a set of $65$ values about the 5 closest opponents to the goal, $13$ values for each opponent including relative position to the goalkeeper and velocities; and \textit{teammate features}, a set of $12$ values about the 4 closest teammates to the goal, $3$ values for each teammate including relative position to the goalkeeper. Our preliminary experiments showed that adding more than 5 opponents and 4 teammates to the input of the agent does not significantly influence the overall performance. Appendix~\ref{app:states_and_actions} provides a detailed description of the action and state spaces.

\subsection{Reward function} 
In this section, we describe the methodology used for finding a suitable reward function from exchanges with domain experts. More details about the resulting function are reported in Appendix~\ref{app:reward_function}. With the help of a professional goalkeeper, we first defined situations that real human goalkeepers face during training. We translated these situations into training scenarios. From these scenarios, we derived the reward function by asking the expert how a real goalkeeper would act. The simplicity of the scenarios allowed us to quickly train and test the agent's behavior in this first iteration. During this phase, we focused on the qualitative behavior rather than the agent's overall performance. After extensive experimentation, we defined the reward function with three main components: a sparse component for saving goals, a dense component for encouraging play near the middle line, and penalty components for reducing noisy movements. The main motivation for the latter two components is that human players need to conserve energy, whereas the agent has infinite stamina and can exploit this to move more and save more shots. The scenarios used in this first iteration are part of the first phase in the curriculum, while new and more complex scenarios are added in subsequent phases. However, the same reward function applies across all training phases.

\begin{table*}
  \begin{center}
  \scalebox{0.7}{
  \begin{tabular}{cc}
      \begin{tabular}{lccr}
        \toprule 
        \textbf{Method} & \bfseries\makecell[c]{Training \\ evaluation} & \bfseries\makecell[c]{Quantitative\\ evaluation} & \bfseries\makecell[c]{Expert-authored\\ test suite} \\
                        & \multicolumn{2}{c}{Success Rate} & Completion Ratio \\
        
        \midrule 
        Our method          &           $\mathbf{90.0\% \pm 1.22}$    &          $\mathbf{73.46\% \pm 1.04}$    &          $91.5\% \pm 0.3$ \\
        Built-in AI         &           $82.6\% \pm 1.67$             &          $65.58\% \pm 1.18$             &          $\mathbf{94.0\% \pm 0.0}^*$ \\
        \bottomrule
    \end{tabular}
        &
    \begin{tabular}{lcc}
        \toprule 
         \textbf{Method} & \textbf{\makecell{Goal \\ Conceding Rate $\downarrow$}} & \textbf{\makecell{Ball \\ Saving Rate $\uparrow$}}  \\
        \midrule
        Our method          &           $\mathbf{25.25\%}$             &          $\mathbf{54.12\%}$\\
        Built-in AI         &           $29.10\%$                        &          $48.27\%$\\
        \bottomrule
    \end{tabular} \\
  \end{tabular}
  }
  \caption{\textbf{Results compared to the main baseline, the built-in AI in EA SPORTS FC 25.} \textbf{Left}: the performance over three benchmark tests, reported over 5 different seeds. The table shows how our method achieves better performance than the built-in AI, while at the same time passing most of the expert-authored tests. \textbf{Right}: our agent and built-in AI facing an experienced human player. The player plays against either our agent or the built-in AI for 400 games. * The built-in AI and the tests in the expert-authored suite are deterministic. }
  \label{tab:quantitative_results} 
  \end{center}
\end{table*}
\begin{table*}
  \begin{center}
  \scalebox{0.8}{
      \begin{tabular}{lcccr}
        \toprule 
                                & \textbf{Main Agent}   & \textbf{Tuned Agent 1} & \textbf{Tuned Agent 2} & \textbf{Tuned Agent 3} \\
                                & \multicolumn{4}{c}{Success Rate} \\
        \midrule
        Last curr phase         &   $90.0\% \pm 1.22$   &   $88.0\% \pm 2.45$   &   $88.4\% \pm 2.40$   &   $88.6\% \pm 1.67$ \\
        Failed scenario 1       &   $28.4\% \pm 2.88$   &   $77.6\% \pm 1.51$   &   $70.4\% \pm 3.65$   &   $69.0\% \pm 2.24$ \\
        Failed scenario 2       &   $13.8\% \pm 1.64$   &   $00.1\% \pm 0.16$   &   $60.4\% \pm 2.07$   &   $49.2\% \pm 0.83$ \\
        Failed scenario 3       &   $00.0\% \pm 0.00$   &   $00.0\% \pm 0.00$   &   $00.1\% \pm 0.18$   &   $14.0\% \pm 1.00$ \\
        \midrule
        \textbf{Average}        &   $33.0\% \pm 1.15$   &   $41.4\% \pm 1.03$   &   $54.8\% \pm 2.07$   &   $\mathbf{55.2\% \pm 1.43}$ \\
        \bottomrule
      \end{tabular}
  }
  \caption{\textbf{Results of fine-tuning.} We report the success rate of the Main Agent, which is trained from scratch using the main method, and Tuned Agents, which have been fine-tuned sequentially on one failed scenario at a time, evaluated over the last curriculum phase and failed scenarios. Failed scenarios are those identified by domain experts where the main agent under-performed. The last row shows the average result over all the evaluation tests. For each fine-tuning iteration, we perform 200,000 training steps compared to the 600,000 steps in the main procedure. Note that the \textit{``Failed scenario 3''} is inherently challenging for the agent and the success rate is close to the expected upper bound of performance. More details in Section~\ref{sec:quant_exp}.\looseness=-1}
  \label{tab:fine_tuning} 
  \end{center}
\end{table*}

\subsection{Evaluation framework}
\label{sec:evaluation}
We define an evaluation framework composed of three steps. We run the evaluation framework after reward convergence during the main training procedure.

\minisection{Automatic quantitative evaluation.} For evaluating the performance of the agent, we use a specific scenario built similarly to the training ones. This scenario covers multiple situations where the agent faces an opponent that shoots towards the goal for 2,000 shots, using different types of shots with varying levels of difficulty. The metric we use is the percentage of saves -- the higher, the better. This specific scenario is not used for training the agent, but only for evaluating it. Appendix~\ref{app:shooting_scenario} shows details of the scenario.

\minisection{Human qualitative evaluation.} Integrating the agent into the game while keeping the experience enjoyable requires a subjective evaluation from human players, making their feedback crucial for evaluation. To achieve this, we leverage the Quality Verification (QV) team within the game studio, and ask professional players to play the game and provide feedback on how to improve the agent. This is important in scenarios where it performs poorly due to a lack of experience during training.

\minisection{Expert-authored test suite.} To control specific cases -- such as critical, edge or challenging cases -- we deploy a set of qualitative tests. Each test includes a situation (including starting positions, velocities, and rotation of both players and the ball) and hand-crafted conditions to assess whether the behavior of the model meets the designers' expectations. This allows us to keep track of each critical situation automatically. The initial test suite was made for the built-in AI. During manual evaluation, each time a new situation is identified as problematic or failing, a test case is created. 
Keeping the test in the evaluation suite ensures that the behavior stays good even after re-training. A total of 344 tests are created to cover all critical situations. We give examples of such test cases in Appendix~\ref{app:eval_scenarios}.

\subsection{Improving Behavior Through Expert Feedback} 
\label{sec:fine_tuning}
We propose a new framework for incorporating domain experts' feedback to improve performance. After the main training converges, domain experts such as professional goalkeepers and QV testers evaluate the agent's behavior using the evaluation framework described in Section~\ref{sec:evaluation}. Whenever they identify a situation in which the agent does not perform as intended, such as a scenario not covered in the initial training or a discovered exploit in the agent's behavior, they create new scenarios and fine-tune the agent specifically on those. For our fine-tuning process, we re-purpose Replay across Experiment (RaE), a symmetric sampling scheme to leverage offline data from previous experiments to improve exploration and bootstrap learning~\citep{acrossexp}. While the paper shows that this can improve performance when mixed with a majority of online data, we show that this approach can be used with a minority of online data in the sense of a fine-tuning approach. We define $D_{\pi_{0}}$ as the replay buffer collected during the main training process. We then follow these steps:   
\begin{enumerate}
    \item Create a training scenario based \emph{on only one failed scenario} identified during evaluation;
    \item Run a training process using new policy and action-value function networks, and only the new training scenario rather than those defined in the curriculum;
    \item During network updates, for each training batch, sample 50\% of the data from the previous buffer $D_{\pi_i}$ and 50\% from the new online buffer $D_{\pi_{i + 1}}$. Here, $\pi_i$ is the policy trained at previous iteration $i$, while $\pi_{i + 1}$ is the policy trained during fine tuning.
    \item Combine the two buffers after training, so that $D_{\pi_{i + 1}} = D_{\pi_i} \cup D_{\pi_{i + 1}}$. Return to step 1.
\end{enumerate}
This process can be repeated for each failed scenario that testers find. Using this technique, we can leverage the knowledge learned from the previous buffer while acquiring new skills to solve the failed scenarios in a sample-efficient manner, as we will show in Section~\ref{sec:experiments}. Figure~\ref{fig:method} shows an overview of all the components of the algorithm.

\section{Experiments}
\label{sec:experiments}
In this section, we present the experimental results of our method using \textit{EA SPORTS FC 25} as a training and evaluation platform, and in particular the goalkeeper positioning system as control problem (described in Section~\ref{sec:environment}).  Additionally, we provide results using the MuJoCo suite in Appendix~\ref{app:mujoco}, demonstrating how our method generalizes to other environments as well. We conduct three types of studies: a quantitative study, where we test the performance of our agent against the built-in AI; a qualitative study, where we compare the qualitative behavior of our agent to that of the built-in AI; and an efficiency study, where we define ablated versions of our method to assess the contribution of each component to performance efficiency. We train each of our main agents for 600,000 training steps, while for the fine-tuning experiments we retrain the agent for 200,000 steps. On average, it takes between 18 and 24 hours to train the agent using all curriculum phases, plus an additional 3 to 6 hours for each fine-tuning iteration.  More details about the computational resources, network architectures, and all hyper-parameters are provided in Appendix~\ref{app:setup}. 

\begin{figure*}  
    \begin{center}  
        \scalebox{0.95}{  
            \begin{tabular}{ccc}  
                \multicolumn{2}{c}{\includegraphics[width=0.55\columnwidth]{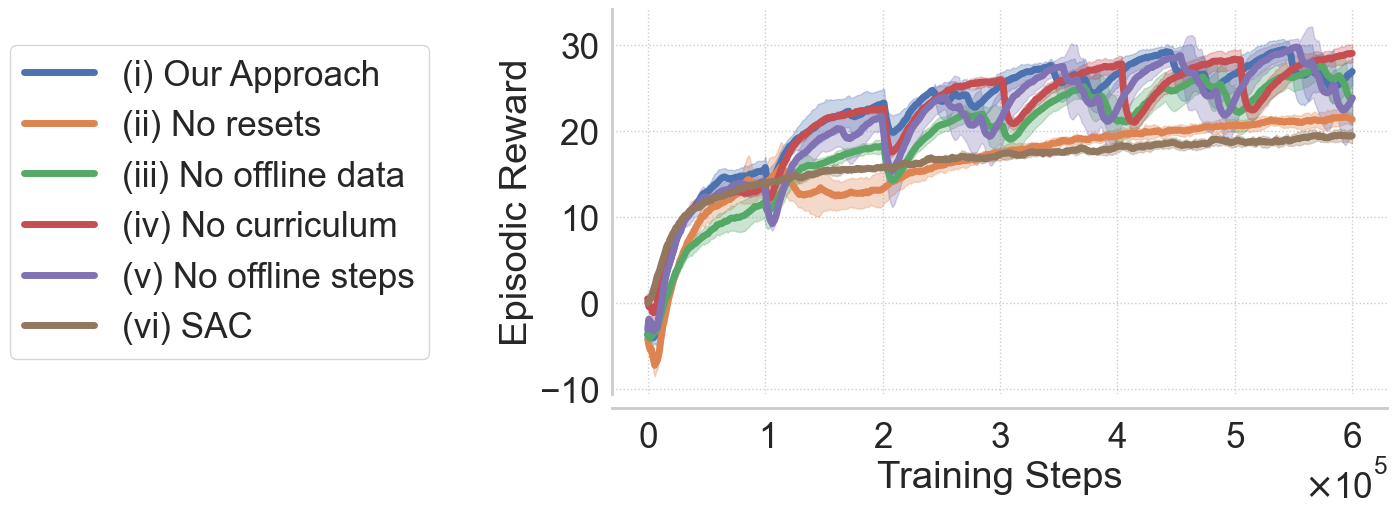}} &  
                \includegraphics[width=0.46\columnwidth]{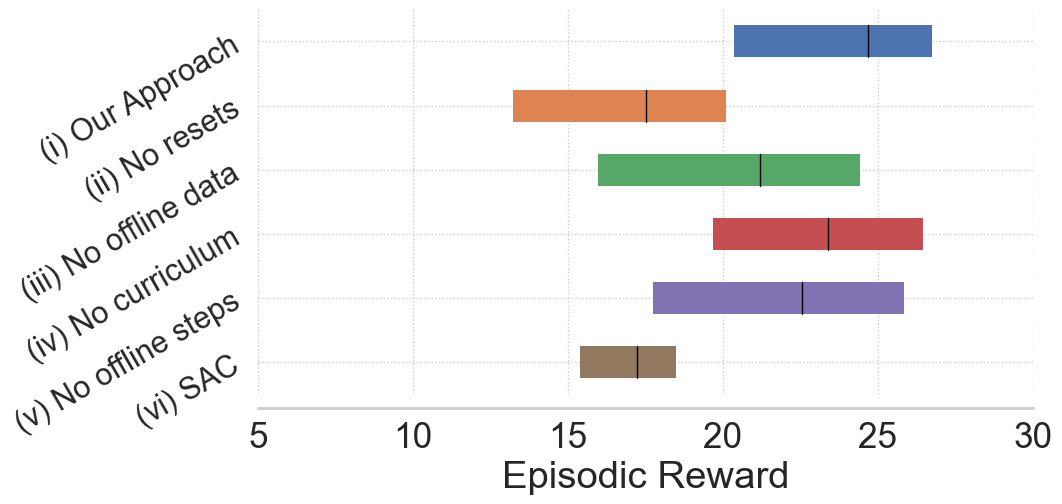} \\  
            \end{tabular}  
        }  
    \end{center}  
    \caption{\textbf{The impact of all the components of our method}. \textbf{Left}: training curves comparing our approach removing different components. The curve represents the mean while the shaded areas represent the standard deviation of 5 different seeds.  The drops in the plot correspond to network resets. \textbf{Right}: interquantile mean of the reward of all the ablations measured at the end of the training. As the plot shows, our agent achieves the highest performance in fewer training steps than the ablations, being more stable.}  
    \label{fig:ablations}  
\end{figure*}  

\subsection{Quantitative study}
\label{sec:quant_exp}
First, we evaluate the performance of an agent trained from scratch using the approach detailed in Section~\ref{sec:algorithm}. Then, we evaluate the performance of our fine-tuning approach, detailed in Section~\ref{sec:fine_tuning}. Our primary baseline is the built-in AI. The main metrics we use are: the mean success rate over 500 episodes using training scenarios; the results of the automatic quantitative evaluation; and the completion ratio of the expert test suite. We repeat each experiment with 5 different seeds.

\minisection{Main training.} We run three benchmarks after training has converged, and we compute: (i) the success rate of a training evaluation benchmark where we test the agent using scenarios in the last phase of curriculum; (ii) the success rate of the automatic quantitative evaluation; and (iii) the completion ratio of the expert-authored test suite. As Table~\ref{tab:quantitative_results} (left) shows, our agent outperforms or is on par with the built-in AI in all tests. Our agent achieves higher success rates in the training scenarios. Additionally, our agent surpasses the built-in AI by almost 10 percentage points in success rate during the quantitative evaluation. In the expert test suite, our agent has similar but lower performance than the built-in AI. This is expected, as most cases in the benchmark were hand-designed with the built-in AI system in mind. However, it is not trivial to tune a machine learning agent to pass these tests. These experiments demonstrate that our reward function produces agents with similar qualitative behavior to the built-in AI but with higher performance. This meets the requirements of developing a game AI that outperforms traditional methods while exhibiting human-like behavior. We want to emphasize that (ii) is the \emph{most meaningful evaluation} because (i) is biased towards our agent, as it is trained to maximize the reward in those scenarios, and (iii) is biased towards the built-in AI, since the tests were originally made for the built-in AI. In contrast, evaluation (ii) is completely independent of both the training and the development of the built-in AI.

\minisection{Fine-tuning.} We deploy our fine-tuning approach on three failed scenarios. Among the failed scenarios, we select three that were especially challenging for the goalkeeper. We first identify the best seed in terms of success rate of the quantitative evaluation resulted from main training process. Then, we evaluate it in the last phase of curriculum and the three failed scenarios. We then sequentially fine-tune the agent on these failed situations. After each step, we evaluate the agent on each of the three failed scenarios as well as the last curriculum phase, collecting the success rate over 500 episodes. 
As Table~\ref{tab:fine_tuning} shows, our approach improves the overall performance in the failed scenarios with fewer resources than the main training. The last agent performs better on average over all scenarios than all other agents. For each iteration, we perform 200,000 training steps compared to the 600,000 steps in the main training. However, after each iteration, the performance in previously failed scenarios degrades slightly. In fact, there is a risk the fine-tuned agent overspecializes to the scenario being used. If we, in one of the next iterations, use a failed scenario where the best performance conflicts with the behavior developed in previous iterations, we may observe a performance degradation. The degradation is not significant in our use case, as the average performance still improves.

{\setlength{\tabcolsep}{1pt}
\begin{figure}
    \begin{center}
    \begin{tabular}{cc}
        \includegraphics[width=0.40\linewidth]{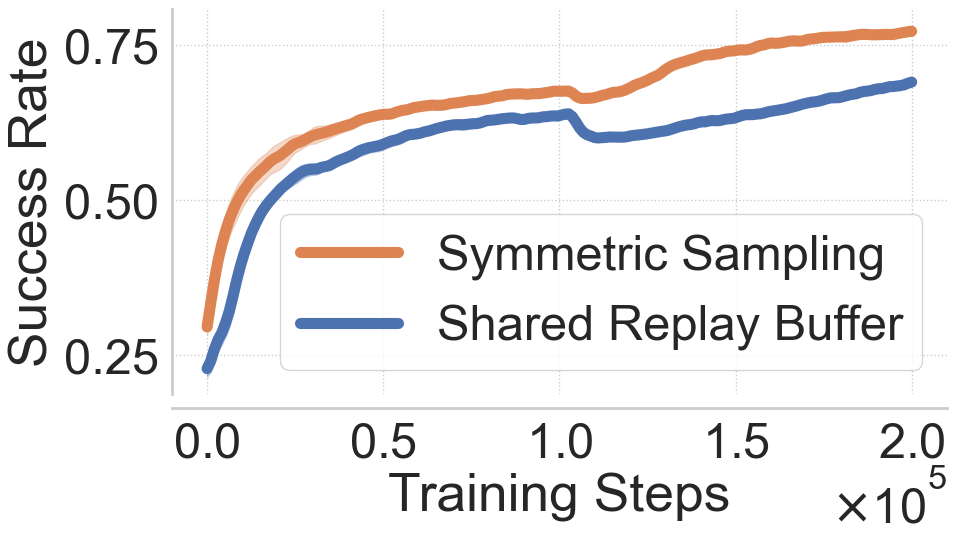} &
        \includegraphics[width=0.40\linewidth]{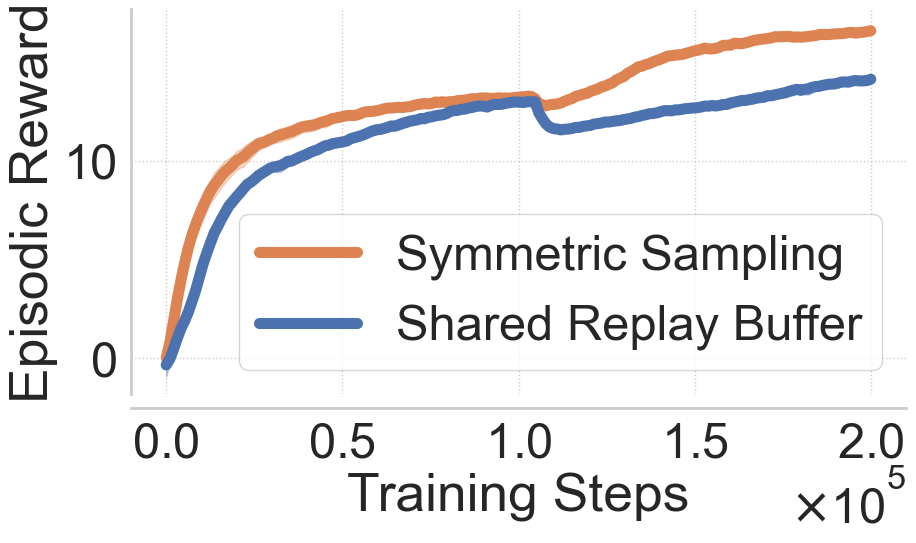} \\
    \end{tabular}
    \end{center}
    \caption{\textbf{The impact of symmetric sampling during fine-tuning}. We compare fine-tuning using the symmetric sampling outlined in Section~\ref{sec:fine_tuning} to a standard approach adding new data to the original buffer. \textbf{Left}: the success rate achieved by both agents in the failed scenario. \textbf{Right}: the episodic reward. Without symmetric sampling, the agent is not able to reach the same performance of our method in the same number of training steps.}
    \label{fig:ablations_ft}
\end{figure}}

\subsection{Qualitative study} 
\label{sec:qual_exp}
We gather feedback from playtesters and expert goalkeepers who evaluate the agent. Moreover, we qualitatively compare our agent's behavior to that of the built-in AI.

\minisection{Human evaluation.} We let an experienced human player compete against our agent and the built-in AI. The human controls one team in a 7v7 scenario. The human player plays 400 games against either our agent or the built-in AI, with the agent type randomly selected at the beginning of each game, which is kept hidden from the player. Each game concludes when the ball goes out of bounds, the goalkeeper catches the ball, or a goal is scored. Table~\ref{tab:quantitative_results} (right) shows the agents' goal conceding rate and the percentage of saves.

Our agent demonstrates higher performance than the built-in AI, even against an experienced human player. An interesting finding is the percentage of saves: our agent is significantly more capable of saving and catching shots from the human player. The human playtester states: \textit{``The movement of the goalkeeper is much more realistic and, from an end-user point of view, is much more enjoyable to play against/with. The goalkeeper is much more reliable, and when one manages to score, it is a very rewarding feeling.''} Furthermore, their personal favorite aspect of the agent is during \textit{``one versus one situations or passes across the box, where these chances are no longer guaranteed goals.''} Finally, they note, \textit{``How the goalkeeper approaches these scenarios is perfect, only jumping on the ball when it is loose from the attacker's feet.''}


\minisection{Behavior analysis.} Figure~\ref{fig:results_summary} compares the behavior of our agent and the built-in AI in one scenario. The figure shows how our agent better understands the situation, being more proactive and trying to anticipate the strikers before they shoot. This is a typical human-like behavior one can see in real football matches. In contrast, the built-in AI is more passive and waits for strikers, allowing them to score goals. We present an extended version in Appendix~\ref{app:behavioral_analysis}.

\subsection{Ablation studies}
\label{sec:ablations}
To evaluate the efficiency of our method compared to standard approaches, we define several ablated versions. As we mentioned in Section~\ref{sec:algorithm}, our primary baseline is standard SAC, as it is the state-of-the-art approach for video game researchers~\cite{sophy,alonso2020deep,bairamian2023minimax}. A key consideration is the data-collection bottleneck. On a standard development machine (e.g. Nvidia RTX 4090), our environment allows for a throughput of ~120 samples per second, compared to other modern RL environments and benchmarks achieving thousands of samples per second\cite{suarez2025pufferlib}.

\minisection{Main training.}
We analyze the impact of each component detailed in Section~\ref{sec:algorithmology} by removing them from our method. Figure~\ref{fig:ablations} shows the results, highlighting the performance of: (i) our approach, (ii) our approach without high replay ratio and resets, (iii) without leveraging offline data, (iv) without curriculum learning using only the final phase of the curriculum, (v) without offline steps during resets, and (vi) using standard SAC. The results demonstrate that the component removed in (ii) is fundamental to the agent's performance. Incorporating resets increases the plasticity of the networks and prevents them from becoming stuck in local optima~\citep{srspr}. Additionally, as shown in the training curves in Figure~\ref{fig:ablations}, offline data (iii) significantly improves efficiency, particularly in the initial phases of training. While (iii) achieves similar performance asymptotically, the performance gap at the beginning is evident.
The figure shows that variation (iv) performs similarly to our agent. However, our method reaches the highest performance peak more quickly and is more stable: the standard deviation of (iv) is higher than that of (i). This can be attributed to the inclusion of all easier scenarios from previous phases in the final phase of the curriculum used by (iv). Without proper complexity scaling, training becomes noisy, increasing result variability and reducing stability. The same applies for (v): while the average performance is similar to our agent, without offline steps the training is more unstable. The difference can be seen in the figure: while (v) can sometimes outperform (i), the standard deviation of (v) is a sign of reduced training stability. To improve stability, it therefore makes sense to introduce offline steps.

\minisection{Fine-tuning.} We analyze the impact of the fine-tuning approach described in Section~\ref{sec:fine_tuning} by removing the symmetric sampling. For this ablation we repeat the fine-tuning experiment described in Section~\ref{sec:quant_exp}, but only for the failed scenario 1. We continue training the main agent using only the failed scenario, but adding the experience to the same buffer created during the main training $D_{\pi_0}$. We do not use symmetric sampling, but we sample randomly from the shared buffer. We start with randomly initialized policy and action-value function networks, and we keep the same reset scheduler. We train the agents for 200,000 steps. Figure~\ref{fig:ablations_ft} shows that the agent with symmetric sampling outperforms the ablated agent in both performance and training speed. 

\section{Limitations and future work}
Although the proposed fine-tuning method enables simple adjustments to the policy, our evaluation shows that repeated fine-tuning leads to a loss of performance on earlier scenarios. This could be caused not only by the loss of plasticity, but also by catastrophic forgetting~\citep{sutton_continual}. This highlights a fundamental trade-off between plasticity and stability, suggesting future work on automated mechanisms for balancing targeted adaptation with retention.

In DRL for games, it is common to train agents against an already existing built-in AI or with self-play. However, to cover the range of situations that can occur in matches with humans, we believe training with human data could bring many advantages. However, it is unclear how to separate and leverage data coming from different types of players, and how to successfully learn from this data. 

We aim to balance strong quantitative performance with realistic qualitative behavior.

In settings such as robotics, autonomous driving, and especially video games, we aim to balance strong quantitative performance with realistic qualitative behavior. Our evaluations show that our agent exhibits noisier behavior compared to the built-in AI. Although this issue does not affect the final performance, we believe there is room for improvement to train smoother policies.

\section{Conclusions}
We presented a sample-efficient DRL method able to train human-like AI. Our method trains agents more than 50\% faster than using standard DRL algorithms. Our ablations show the contribution of our method to the general sample efficiency landscape in DRL. To prove the practicality of the method, we tested it in \textit{EA SPORTS FC 25}, a commercial football video game. By leveraging and improving on the latest advancements in sample-efficient DRL, our method can train well-performing agents in less than one day. Moreover, our approach allows game developers to adjust the policy's performance in case of bad behaviors without retraining it from scratch. Our agent outperforms the existing built-in AI in both quantitative and qualitative terms. 

\bibliographystyle{rlj}
\bibliography{references.bib}
\newpage
\appendix
\onecolumn
\section{Soft Actor-Critic}
\label{app:sac}
We consider a discrete-time Markov Decision Process (MDP), which consists of a tuple $\langle S, A, R, P, \gamma \rangle$. All the elements in the MDP are defined in Section~\ref{sec:preliminaries}. Soft Actor-Critic (SAC) focuses on the maximum entropy reinforcement learning setting, where the agent's objective is to find the optimal policy $\pi^*$ which maximizes the expected cumulative reward while keeping the entropy $\mathcal{H}$ of the policy distribution high:
\begin{equation}
    J = \arg\max_{\pi^*} \mathbb{E}_{s_0 \sim P}\left[\sum_{t=0}^{\infty} \gamma^t(r_t - \alpha \mathcal{H}(\pi(\cdot \;, s_t)))\right],
\end{equation}
where $\alpha$ is an hyper-parameter. The action value function is defined by:
\begin{equation}
    Q(s, a) = \mathbb{E}\left[\sum_{t=0}^{\infty} \gamma^t(r_t - \alpha \log(\pi(a_t | s_t))) \; | \; s_0 = s, a_0 = a\right].
\end{equation}
$Q(s, a)$ describes the expected future discounted reward gained by taking action $a_t$ at a particular state $s_t$ at timestep $t$. In particular, SAC parametrizes the action value function and policy as neural networks and trains two independent versions of the $Q$ function, using the minimum of their estimates to compute the regression targets for Temporal Difference (TD) learning. The optimization objective for the $Q$ functions is:
\begin{equation}
    L_Q = \mathbb{E}_{(s, a, r, s', d) \sim D}[(Q_i(s, a) - y(r, s', d))^2],
\end{equation}
where $D$ is the dataset, $d$ is a value indicating whether the episode is terminated, $Q_i$ is the action value function with $i=1,2$, and $y$ is defined as the target value:
\begin{equation}
    y(r, s', d) = r + \gamma (1-d) (\min_{i=1,2} Q_{\text{target}_i}(s', a) - \alpha \log \pi(\hat{a}', s'), \;\;\;\; \hat{a}' \sim \pi(\cdot \;, s'),
\end{equation}
where $Q_{\text{target}_i}$ is the target action value function, a copy of $Q_i$ initialized with the same initial weights as $Q_i$.
Finally, the policy is optimized accordingly to:
\begin{equation}
    L_\pi = \max \mathbb{E}[\min_{i=1, 2}Q_i(s,\hat{a}) - \alpha \log \pi(\hat{a}, s)], \;\;\;\; \hat{a} \sim \pi(\cdot, s).
\end{equation}
After each iteration, SAC updates the weights of the target value functions through a soft update:
\begin{equation}
    \phi_{\text{target}_i} \leftarrow \rho \phi_{\text{target}_i} - (1 - \rho)\phi_{i},
\end{equation}

where $\phi_{\text{target}_i}$ is the set of weights for $Q_{\text{target}_i}$, $\phi_{i}$ is the set of weights for $Q_{i}$, and $\rho$ is a hyper-parameter. 

\section{Additional Qualitative Behavior Analysis}
\label{app:behavioral_analysis}
\begin{figure}
    \centering
    \includegraphics[width=\linewidth]{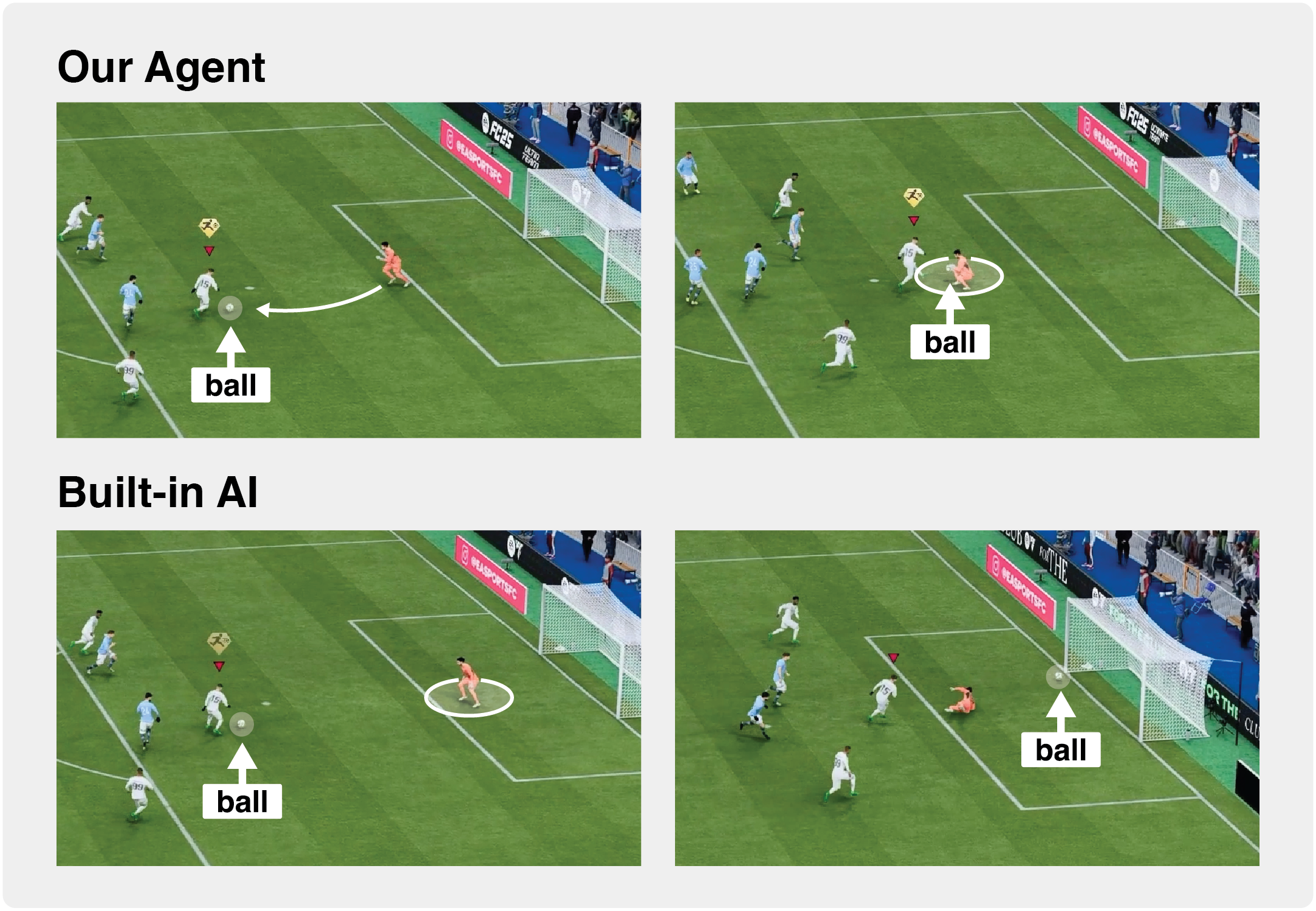}
    \caption{\textbf{An example showcasing the behavioral differences} between our agent (top) and the built-in AI (bottom) in the same situation. Our agent is more proactive and better understands the current situation, anticipating the shot and moving forward early. In contrast, the built-in AI is more passive, allowing the striker to score a goal by leaving the goal cage open. We report a quote from a professional goalkeeper: \textit{``the goalkeeper plays it really well! Keeper looks for opportunities to steal ground as the striker enters the box.''}}
    \label{fig:qual_behavior_1}
\end{figure}

\begin{figure}
    \centering
    \includegraphics[width=\linewidth]{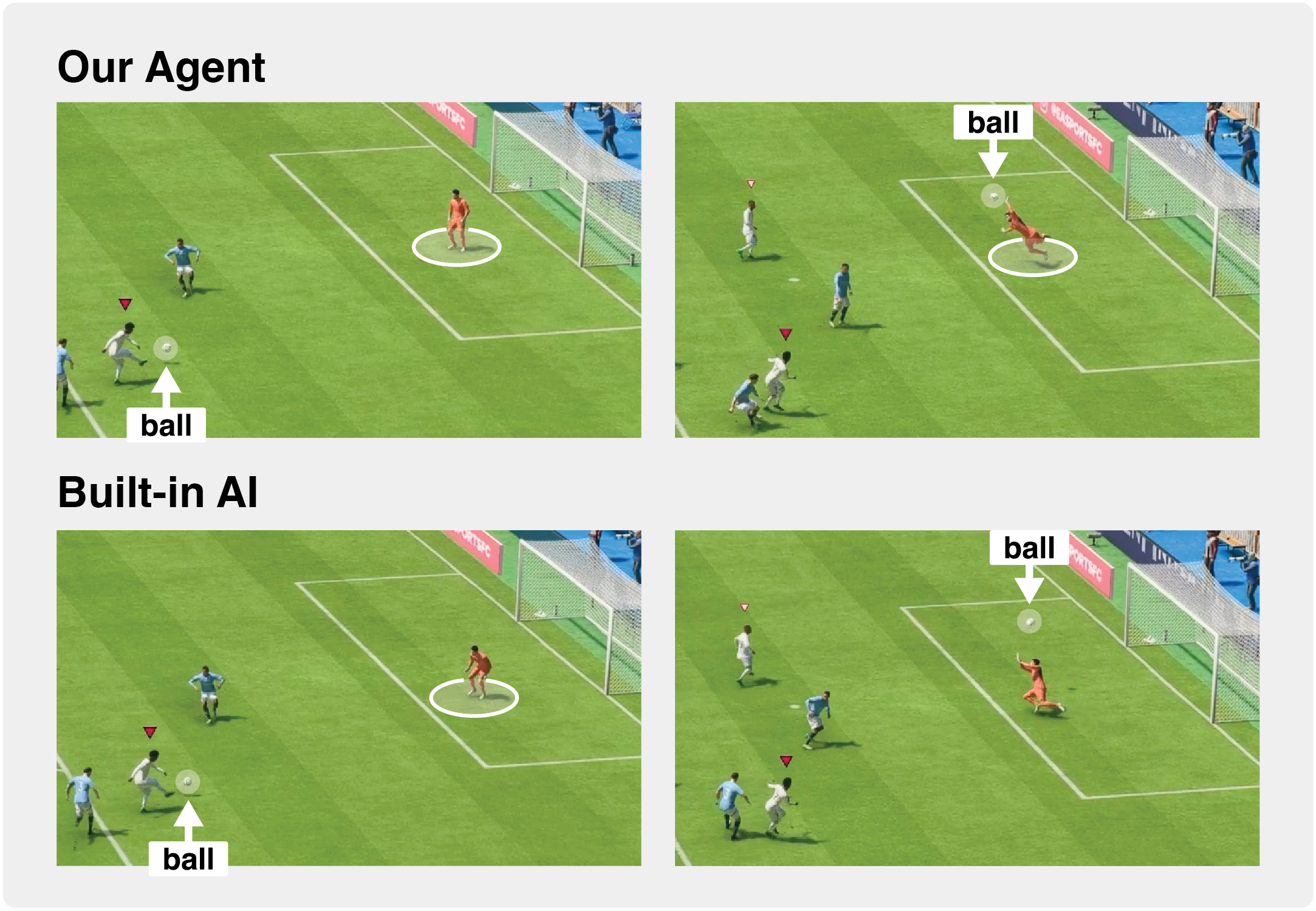}
    \caption{\textbf{An example showcasing the behavioral differences} between our agent (top) and the built-in AI (bottom) in the same situation. Our agent better covers the middle-line --  the line between the ball and the center of the goal -- being able to save the distant shot from the striker. The built-in AI is less patient, covering the first post --  the post closer to the ball -- leaving enough space for the striker to score. We report a quote from a professional goalkeeper: ``\textit{Good positioning by the goalkeeper, and the depth is good}.''}
    \label{fig:qual_behavior_2}
\end{figure}

\begin{figure}
    \centering
    \includegraphics[width=\linewidth]{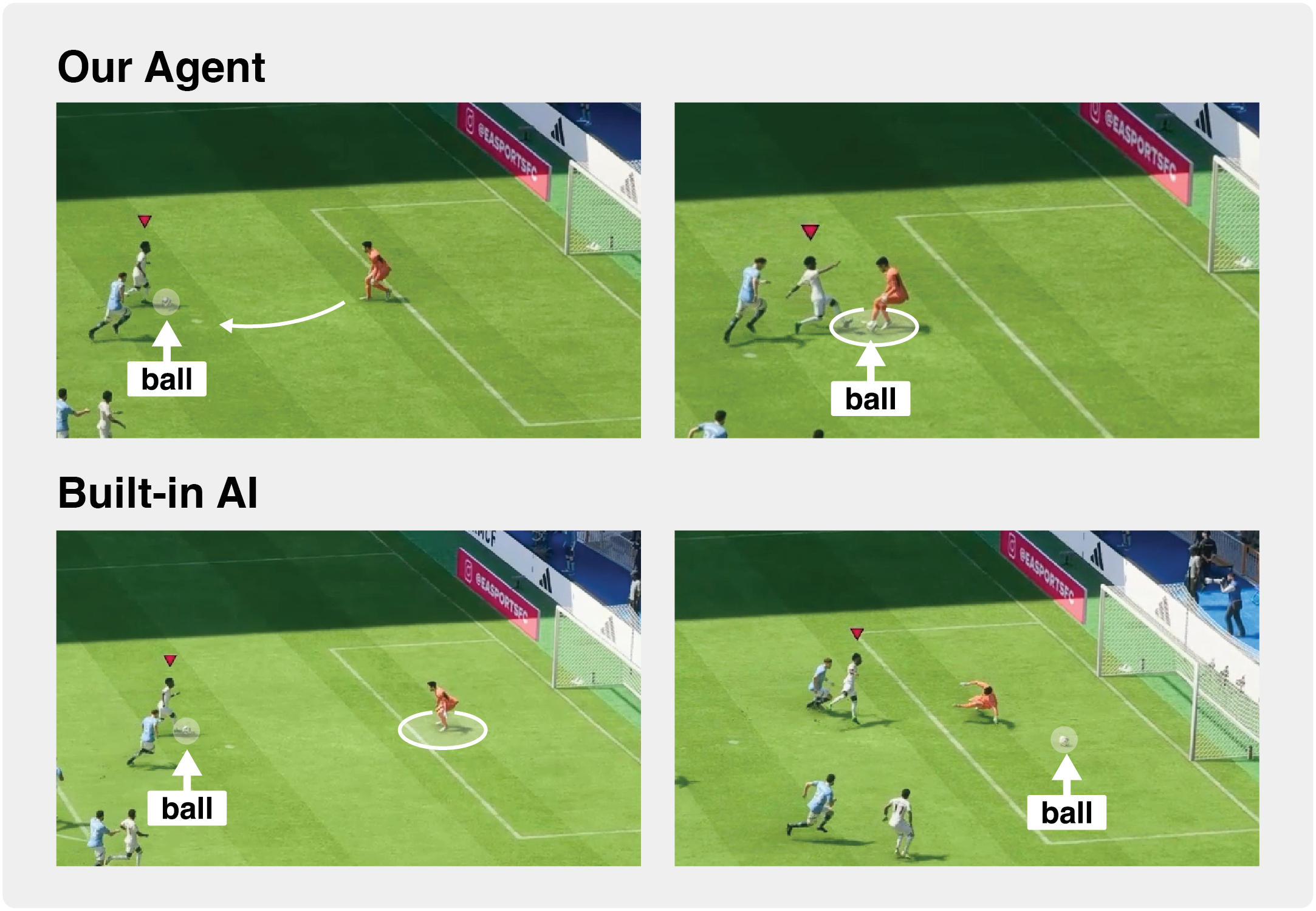}
    \caption{\textbf{An example showcasing the behavioral differences} between our agent (top) and the built-in AI (bottom) in the same situation. Our agent understands that in particular 1v1 situations, in order to increase the probability of saving the ball, it needs to ``close the space'' of the striker. Instead, the built-in AI always use the same passive behavior, giving more chances to the striker.}
    \label{fig:qual_behavior_3}
\end{figure}

\begin{figure}
    \centering
    \includegraphics[width=\linewidth]{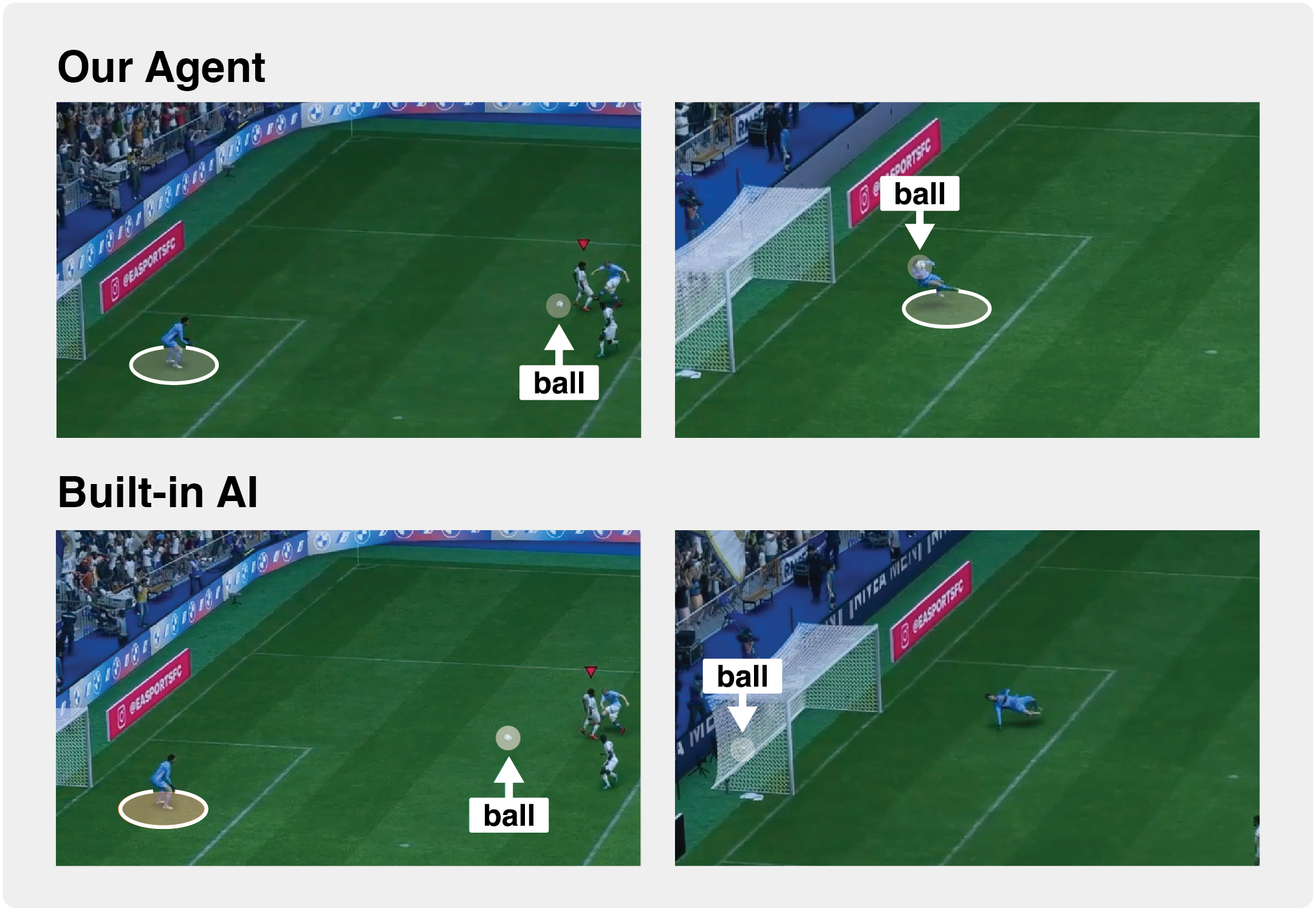}
    \caption{\textbf{An example showcasing the behavioral differences} between our agent (top) and the built-in AI (bottom) in the same situation. The starting position of the agent is very similar to the one of the built-in AI, but it is different enough so that our agent can catch the ball, while the built-in AI can not.}
    \label{fig:qual_behavior_4}
\end{figure}

In this section we expand the qualitative behavior analysis outlined in Section~\ref{sec:qual_exp}. Figure~\ref{fig:qual_behavior_1}, Figure~\ref{fig:qual_behavior_2}, Figure~\ref{fig:qual_behavior_3}, and Figure~\ref{fig:qual_behavior_4} show different scenarios comparing the qualitative behavior of our agent compared to the built-in AI. The results demonstrate, across all the scenarios shown, that our agent performs better than the built-in AI mainly because it exhibits a more human-like behavior than the hand-crafted AI. In Figure~\ref{fig:qual_behavior_1} and Figure~\ref{fig:qual_behavior_3}, the agent exhibits a more proactive behavior, similar to that of a human goalkeeper. According to a professional goalkeeper, in order to increase the probability of catching the ball in a 1v1 situation, a real human goalkeeper should ``close the space'' of the striker. Figure~\ref{fig:qual_behavior_2} shows the agent achieve better positioning by covering the middle-line -- the line between the ball and the center of the goal -- successfully saving a shot from the distance. Instead, the built-in AI leaves space for the striker to shoot to the ``far post.'' In Figure~\ref{fig:qual_behavior_4}, the starting position of the agent is very similar to the one of the built-in AI, but it is different enough so that our agent can catch the ball, while the built-in AI can not.

\clearpage

\clearpage

\clearpage

\clearpage

\section{Action and state spaces}
\label{app:states_and_actions}
The action space of the agent consists of three continuous values: two values for the relative XZ components of the movement vector, and one value for the intensity of the movement. The latter controls how fast the agent should reach the position defined by the movement vector. 

A variety of state features are input to the neural networks. We use the same state space for both the agent and action-value functions. These features are directly available from the game engine. Each of the features is normalized to the $[-1, 1]$ range. We divide the state space into three main components, for a total of 109 features. The first component comprises the set of goalkeeper egocentric information, such as position and velocities; the second component includes a set of opponent features; and the third includes a set of features for the goalkeeper's teammates. The features are engineered to provide the agent a fully observable environment.

\section{Reward Function}
\label{app:reward_function}
The reward function is hand-crafted for achieving a human-like behavior. Several aspects of the reward function were directly influenced by feedback from a professional goalkeeper who evaluated the agent during initial development. Extra attention was given to complex scenarios such as lobs and many-vs-many situations. It is composed of three reward components: a sparse reward for catching or deflecting the ball, a dense reward for encouraging the player to cover most of the area delimited by the goal, and penalties to incentivize smooth movements.   

We argue that leveraging domain expertise is fundamental for developing a DRL agent that exhibits good qualitative behavior suitable for complex practical settings, such as video games. For instance, in the initial iterations, it was relatively easy to train an agent capable of beating the built-in AI, but it exhibited noisy behaviors that would detract from its credibility and the ability to be fun to play against. With the help of domain expertise, we realized that smoother behavior is more important than pure performance.

\section{Training Setup and Hyper-parameters}
\label{app:setup}
\begin{figure}
    \centering
    \begin{tabular}{cc}
         \includegraphics[width=0.46\columnwidth]{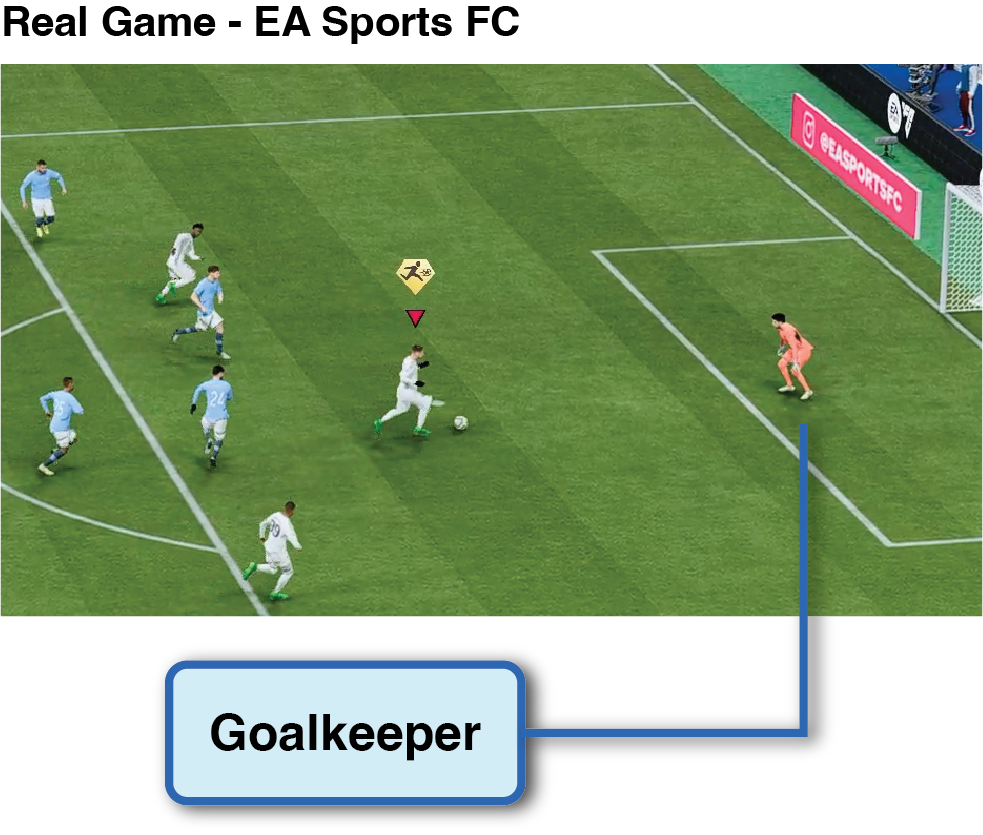}  &
         \includegraphics[width=0.46\columnwidth]{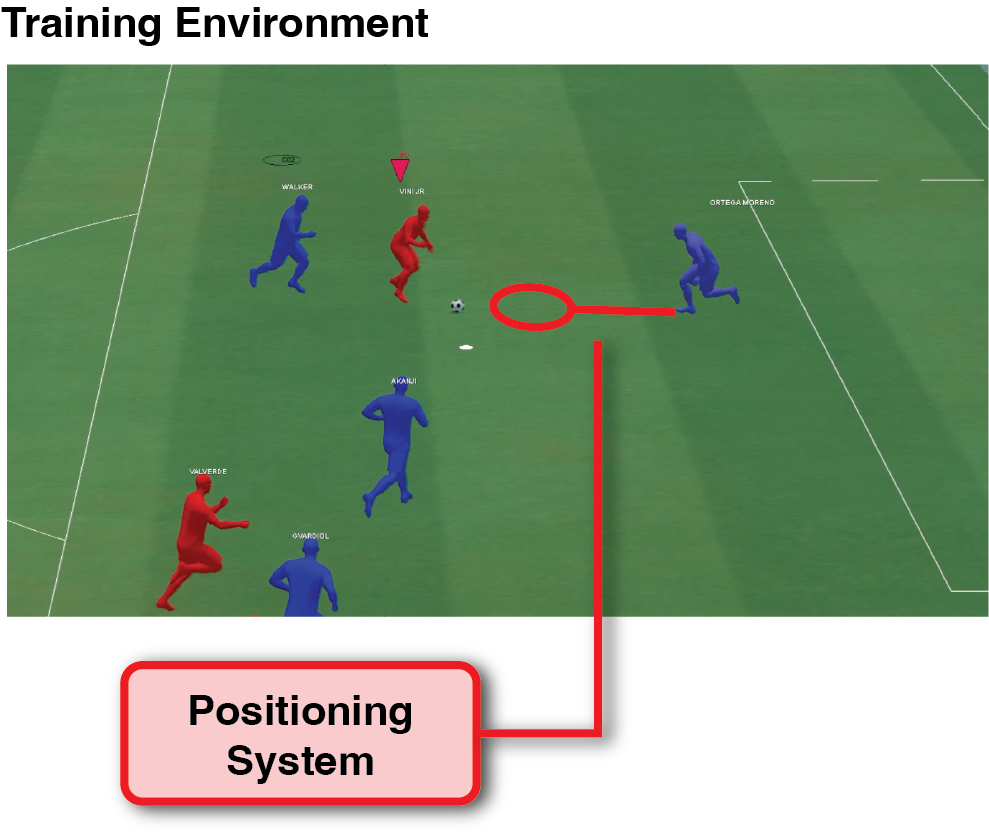} \\
    \end{tabular}
    
    \caption{
    \textbf{Overview of EA SPORTS FC 25 and the goalkeeper positioning system}. \textbf{Left}: screenshot of the real game, \textbf{Right}: screenshot of the training environment. The training environment has the exact same gameplay mechanics (logic, physics, etc.) as the real game, but it lacks all the graphical enhancements. The goalkeeper positioning system outputs a target position, shown by the red line and circle in the right image, and the desired intensity of the movement.
    } 
    \label{fig:fc}
\end{figure}

All training was performed deploying 4 parallel environments on the same machine with an NVIDIA A6000 GPU with 48GB RAM and a AMD Ryzen Threadripper PRO 7975WX 32-Core CPU. As mentioned in Section~\ref{sec:environment}, we train our agent using a low-resolution version of the game, in which we can unlock the frame rate allowing us to speed-up the simulation by a factor of 3. Figure~\ref{fig:fc} shows the difference between the real game and the low-resolution version. 

\subsection{Learning Architecture Details}
As previously mentioned, we use Soft Actor-Critic (SAC)~\citep{sac} as the optimization algorithm. Standard SAC requires a policy network and two identical action-value function networks. We use the same architecture for both the policy and action-value function networks, but we initialize them independently. In order to satisfy the runtime requirements for running the policy network in a game at 60 FPS, we constrain the network's size. Both the policy and the action-value functions are represented by 5-layer MLPs, all layers with a size of 256, ReLU activations, and layer normalization. The last layer of the policy outputs the mean and standard deviation of a diagonal Gaussian distribution for each of the three actions. The action-value functions have a last layer of size 1 without activation, returning the estimated future discounted reward.

\subsection{Curriculum Learning}
\label{app:curriculum}
We let $C^i$ represent the collection of scenarios in phase $i$. To mitigate the risk of catastrophic forgetting, we reuse scenarios encountered in previous phases such that $C^{i} \cap C^{i-1} \neq \emptyset,  \;\; \forall  \;\; 2 \le i \le N$, where $N$ is the number of phases in total. We move from $C^i$ to $C^{i+1}$ when the agent has reached an average success rate in all scenarios of $C^i$. The scenarios retained from the previous phase and their number, as well as the success rate threshold for each phase, are manually determined.

\begin{figure}
    \centering
     \includegraphics[width=0.80\linewidth]{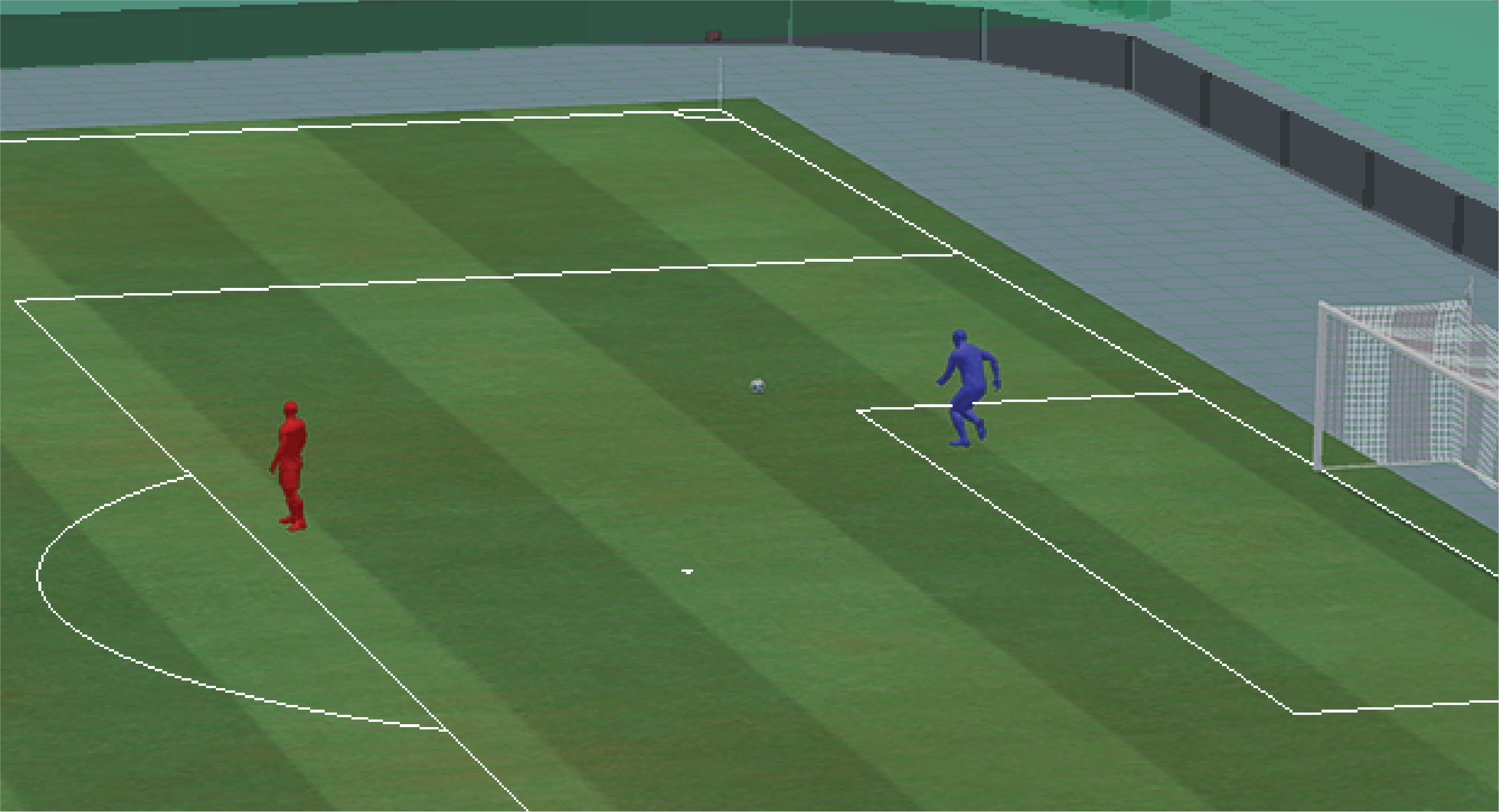}
    \caption{
    \textbf{Example of scenario within phase 1.} In this scenario, the agent (blue) is tasked to catch a ball that is slowly rolling away from the goal cage. There is also a single striker (red) who does not act. This simple situation helps the agent learn what it means to catch the ball at the beginning of the training.} 
    \label{fig:curr_1}
\end{figure}

\begin{figure}
    \centering
     \includegraphics[width=0.80\linewidth]{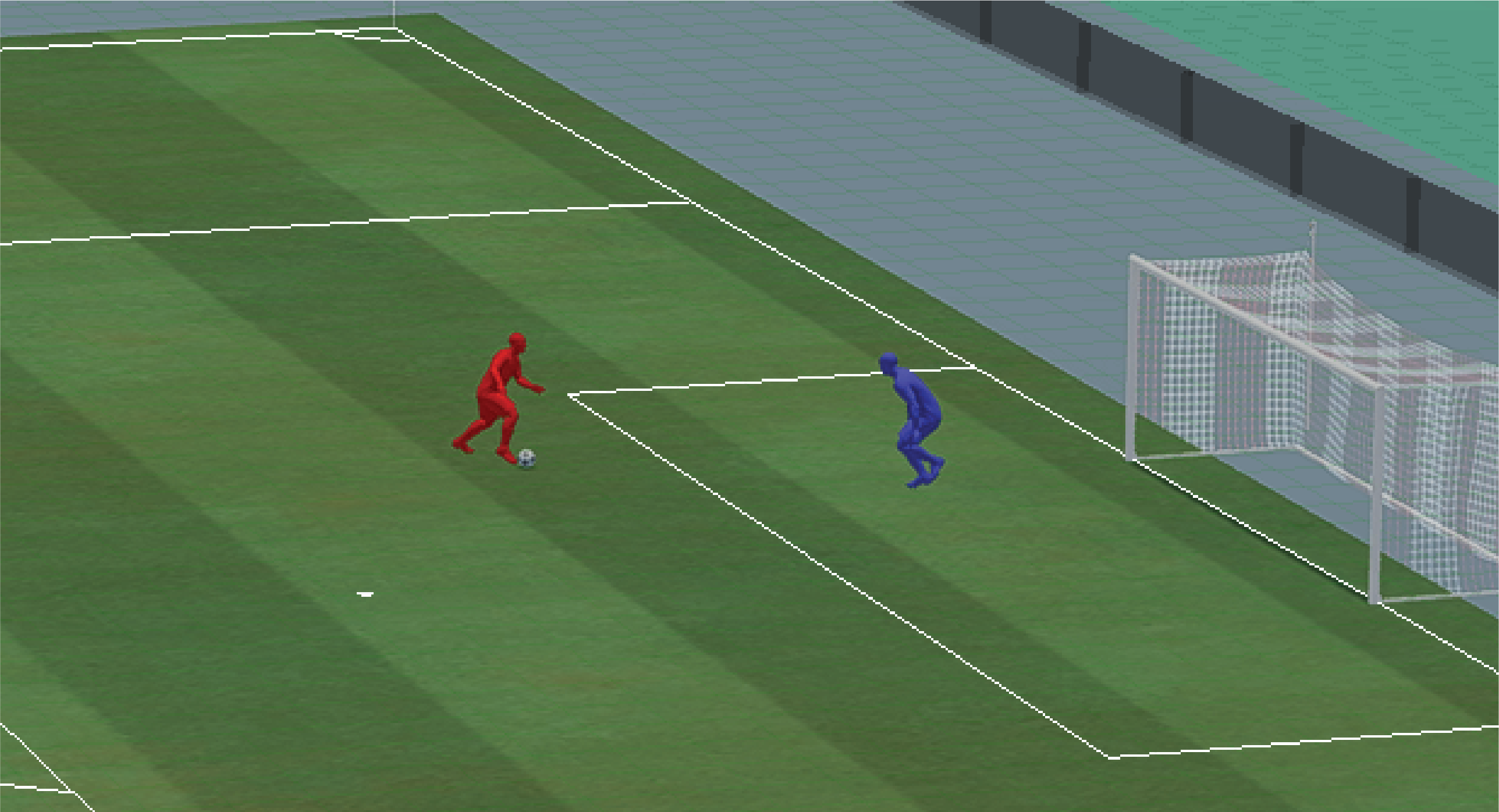}
    \caption{
    \textbf{Example of scenario within phase 1.} In this scenario, we task the agent (blue) to learn a common situation in football: playing 1v1 with a striker (red). Although it may seem simple, this scenario contains hidden complexities depending on the expertise of the striker and the starting positions of players. For this phase, we use a naive striker who moves toward the goal cage and shoots when it sees an opportunity. We maintain this scenario in subsequent phases of curriculum learning. Additionally, more complex versions of this scenario are added in later phases.} 
    \label{fig:curr_2}
\end{figure}

\begin{figure}
    \centering
     \includegraphics[width=0.80\linewidth]{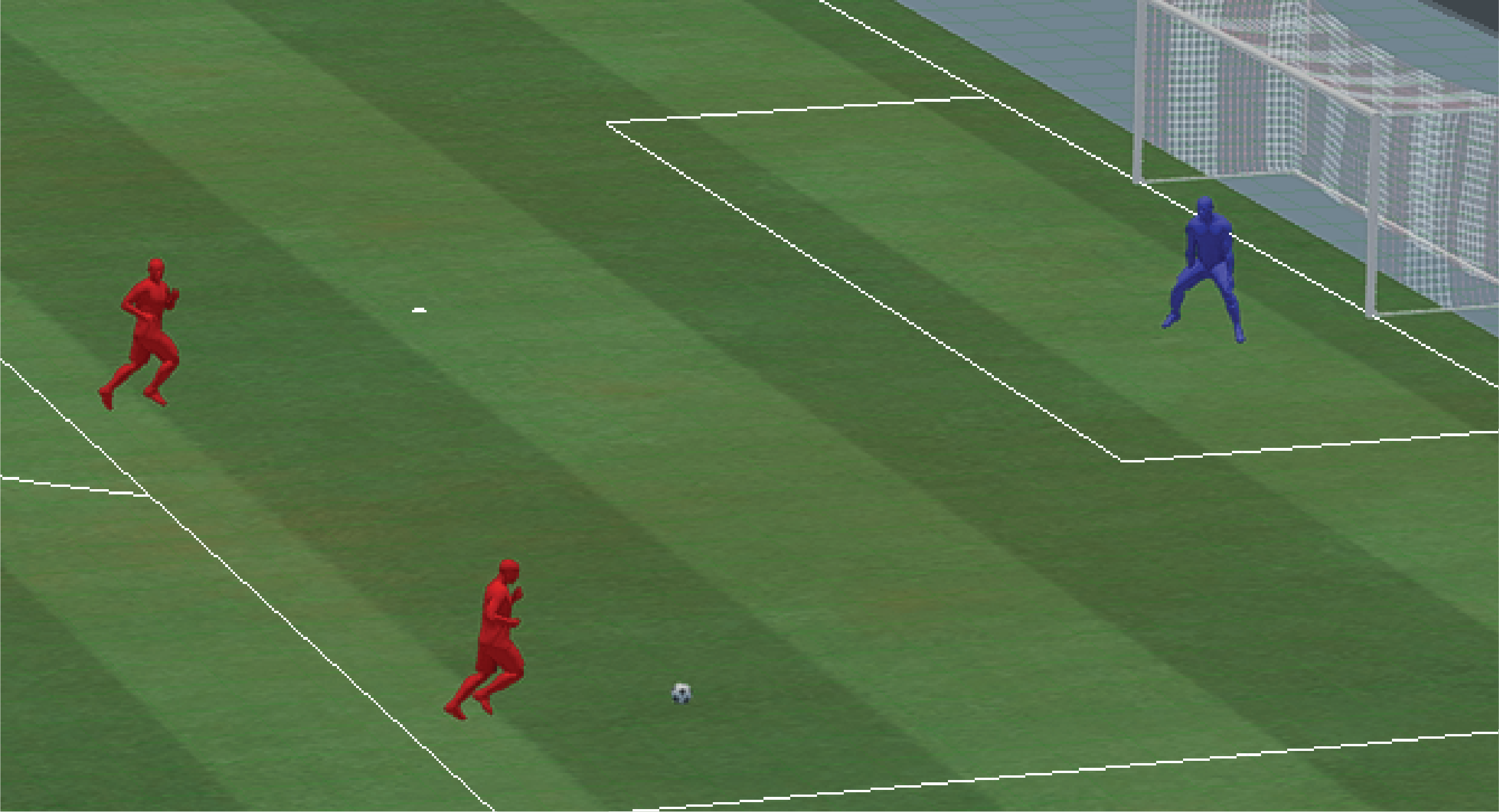}
    \caption{
    \textbf{Example of scenario within phase 2.} In this scenario, we show an evolution of the situation shown in Figure~\ref{fig:curr_1}: a 2v1 scenario. Two strikers (red) face off against the agent (blue). This situation is more complex than the 1v1 scenario because the agent needs to decide whether to rush out towards the strikers if it perceives shooting intentions, or to wait for a pass between the strikers.}
    \label{fig:curr_3}
\end{figure}

\begin{figure}
    \centering
     \includegraphics[width=0.80\linewidth]{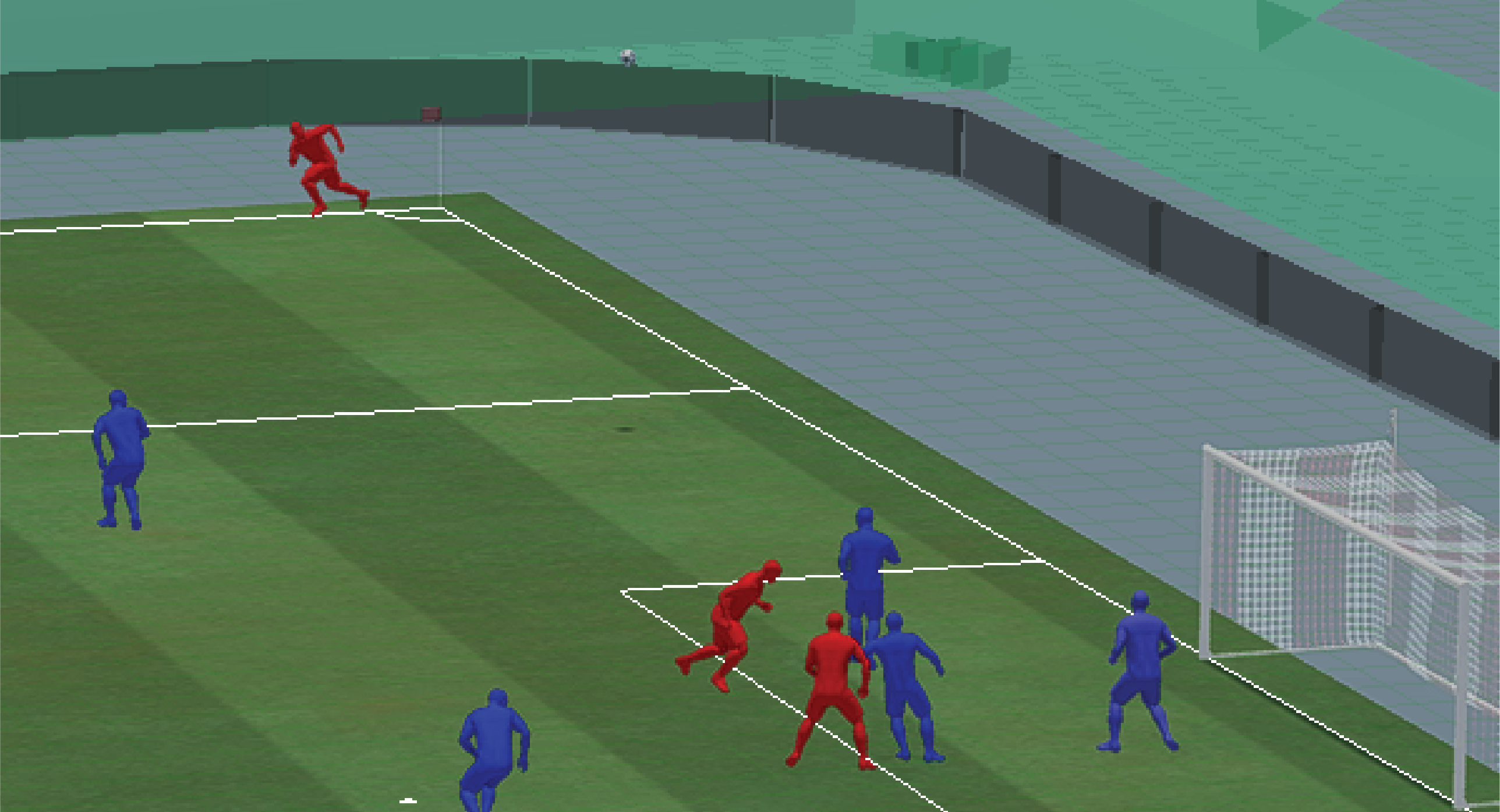}
    \caption{
    \textbf{Example of scenario within phase 2.} In this scenario, we present a common situation in football: a corner kick. The red players are the agent's opponents, while the blue players are the agent's teammates. In this situation, the agent must decide whether to rush out and reach the ball before the other strikers, risking a shot; or to wait and allow the striker to shoot, aiming for a safe catch. This situation shows an example scenario that we keep in phase 3.}
    \label{fig:curr_4}
\end{figure}

\begin{figure}
    \centering
     \includegraphics[width=0.80\linewidth]{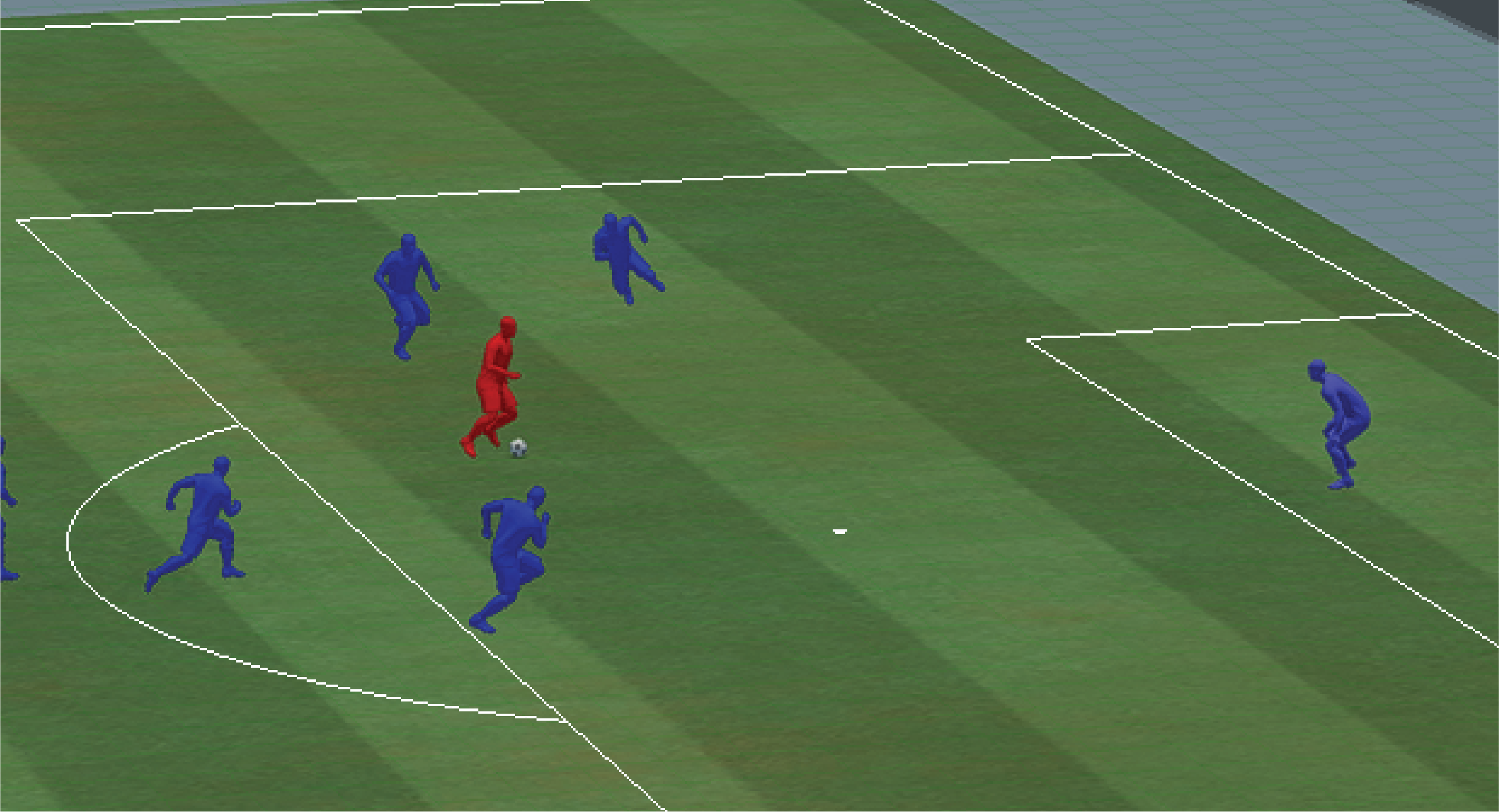}
    \caption{
    \textbf{Example of scenario within phase 3.} 
    The figure shows a general situation: a 7v7 match. In this scenario, all players, including strikers and defenders, do not follow any specific behaviors. We can see a striker (in red), while the blue team represents the agent's teammates. The scenario starts in the agent's half of the pitch and ends if the ball goes out of bounds, the goalkeeper catches the ball, or after a maximum number of steps. We have different versions of this scenario that vary the difficulty level of the strikers AI.}
    \label{fig:curr_5}
\end{figure}

\begin{figure}
    \centering
     \includegraphics[width=0.80\linewidth]{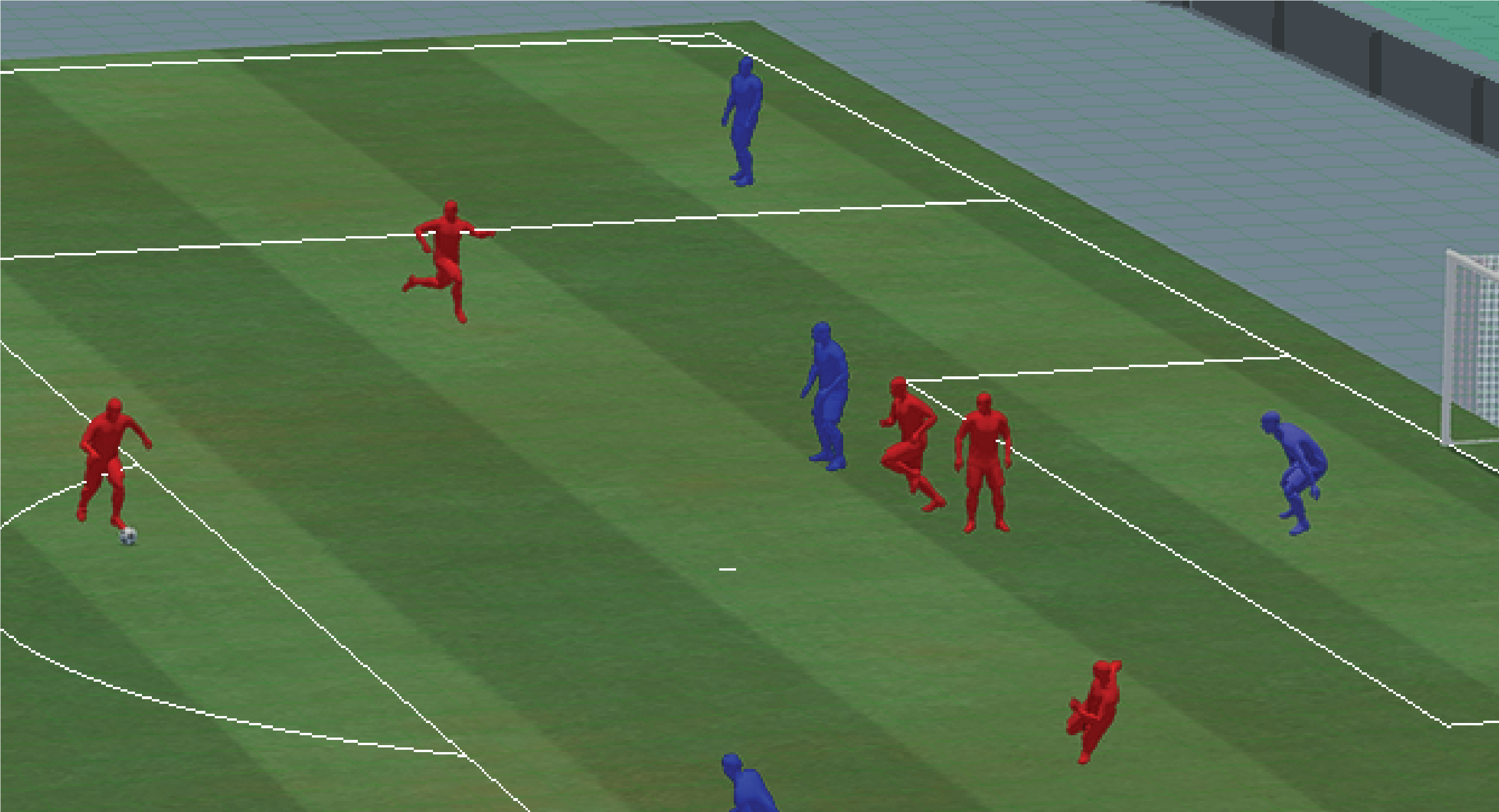}
    \caption{
    \textbf{Example of scenario within phase 3.}
    The figure shows one of the most complex scenarios in all of the curriculum phases. In this situation, 7 strikers (red) with the highest skill-level of AI face 7 defenders (blue) with the lowest skill-level of AI. The goalkeeper cannot rely on its teammates and must correctly understand the situation to find the best position to decrease the probability of a goal.}   
    \label{fig:curr_6}
\end{figure}

Figures from \ref{fig:curr_1} to \ref{fig:curr_6} show examples of scenarios used during curriculum learning. Figures~\ref{fig:curr_1} and~\ref{fig:curr_2} illustrate scenarios from phase 1, Figures~\ref{fig:curr_3} and~\ref{fig:curr_4} from phase 2, and Figures~\ref{fig:curr_5} and~\ref{fig:curr_6} from phase 3. More details of the individual scenarios are provided in the captions. Figures~\ref{fig:curr_1} and~\ref{fig:curr_3} show examples of scenarios that are kept from phase 1 to phase 2 and from phase 2 to phase 3.

\clearpage

\clearpage

\clearpage

\clearpage
\subsection{Hyper-parameters}
Table~\ref{tab:hyper} describes the set of hyper-parameters used in our experiments. 

\renewcommand{\arraystretch}{1.3}
\begin{table*}[h]
  \begin{center}
    \begin{small}
    \scalebox{1.0}{
      \begin{tabular}{l|r}
        \toprule
        \textbf{Parameter} & \textbf{Value} \\
        \midrule
        Online batch size           & $512$ \\
        Offline batch size          & $512$ \\
        Buffer size                 & $10^7$ \\
        Action repetitions          & $5$ \\
        Max timesteps               & $300$ \\
        Discount $\gamma$           & $0.997$ \\
        Optimizer                   & Adam \\
        Learning rate               & $5 \times 10^{-4}$ \\
        $\beta_1$                   & $0.9$ \\
        $\beta_2$                   & $0.999$ \\
        Replay ratio                & $1$ \\
        Number of Curriculum Phases $N$ & $3$ \\
        Reset interval (gradient steps) & $100,000$ \\
        Offline steps               & $6,400$ \\
        Random initial actions      & $25,000$ \\
        Episode of demonstrations for each curriculum phase & $1,000$ \\
        Success rate threshold for each curriculum phase & $[0.90, 0.90]$ \\
        Scenarios in each curriculum phase & $[11, 18, 25]$ \\
        Q networks depth            & $5$ layers \\
        Q networks width             & $256$ \\
        Q networks layer normalization & True \\
        $\pi$ network depth         & $5$ layers \\
        $\pi$ network width         & $256$ \\
        $\pi$ network layer normalization & True \\
        \bottomrule
      \end{tabular}}
      \caption{\textbf{Hyper-parameters.} The most important hyper-parameters of our approach and their respective values.}
     \label{tab:hyper}
    \end{small}
  \end{center}
\end{table*}

\newpage
\section{Evaluation Framework}
\label{app:evaluation}
In this section we describe more details regarding the framework detailed in Section~\ref{sec:evaluation}. Moreover, we show some screenshots of the scenarios used in the benchmarks.

\subsection{Automatic Quantitative Evaluation}
\label{app:shooting_scenario}
\begin{figure}
    \begin{center}
    \begin{tabular}{cc}
        \includegraphics[width=0.48\columnwidth]{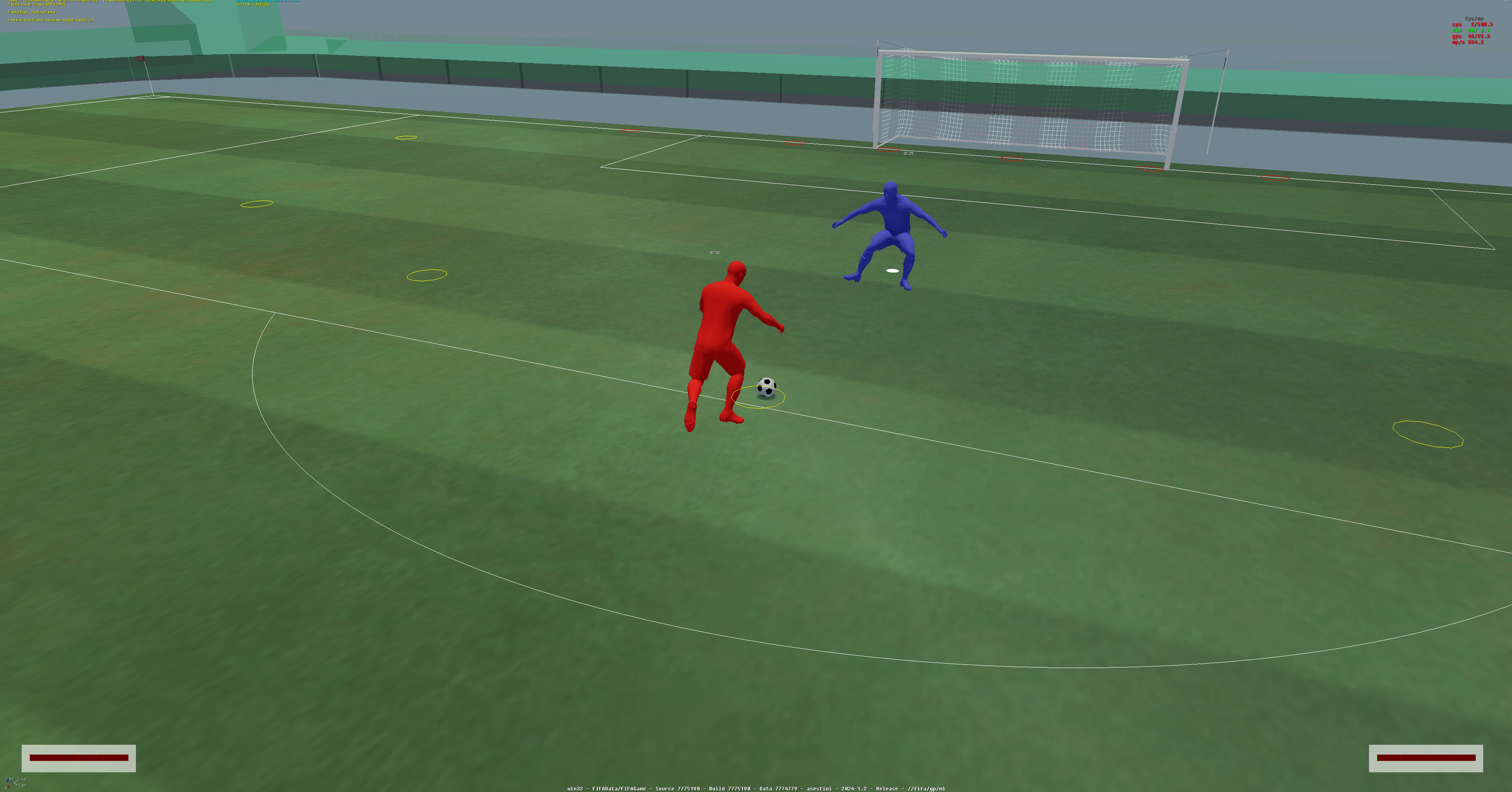} &
        \includegraphics[width=0.48\columnwidth]{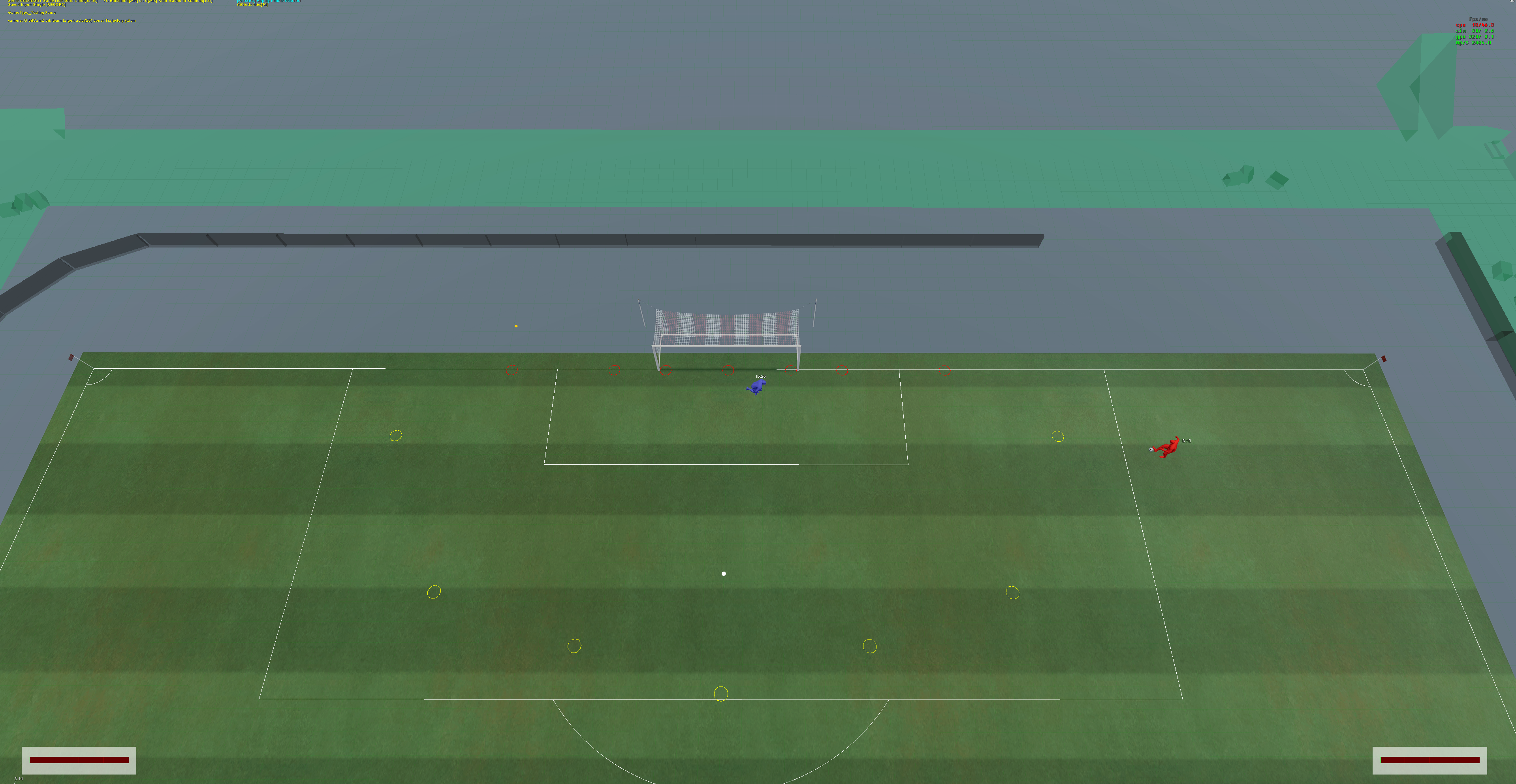} \\
    \end{tabular}
    \end{center}
    \caption{\textbf{Screenshots of the automatic quantitative evaluation}. In the automatic quantitative evaluation, the agent (blue) faces a striker (red) for 2,000 shots. \textbf{Left}: a screenshot of the test running. \textbf{Right}: a top-down view of the test. The figure shows the possible starting position of the striker in yellow, and the possible shooting target in red. For every shot, the striker will randomize the power, the type, and the target of the shot.}
    \label{fig:quant_eval}
\end{figure}

Figure~\ref{fig:quant_eval} shows the scenario used for the automatic quantitative evaluation. It is a simple scenario where the agent faces an opponent that shoots towards the goal for 2,000 steps, using different types of shots with varying levels of difficulty. The possible starting position of the striker is highlighted in yellow, while the possible shooting targets are highlighted in red. 

\subsection{Expert-authored Test Suite}
\label{app:eval_scenarios}
\begin{figure}
    \centering
     \includegraphics[width=0.90\linewidth]{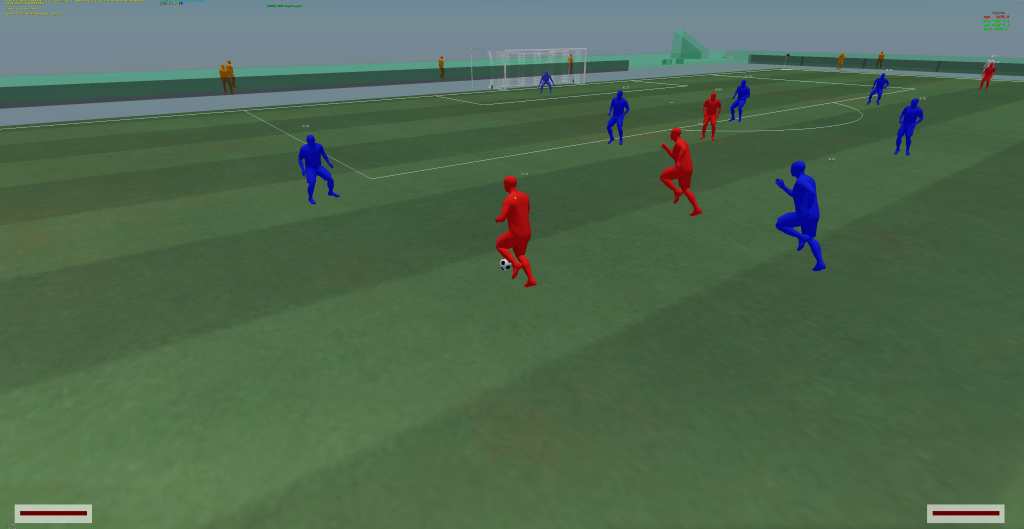}
    \caption{
    \textbf{Example of a test in the expert-authored test suite.} In this test, the striker (in red, in the foreground) is threatening the right post, and human goalkeepers should cover it as fast as they can. For this reason, in this situation the test checks whether the agent is moving towards the right post.}
    \label{fig:qual_eval_1}
\end{figure}

\begin{figure}
    \centering
     \includegraphics[width=0.90\linewidth]{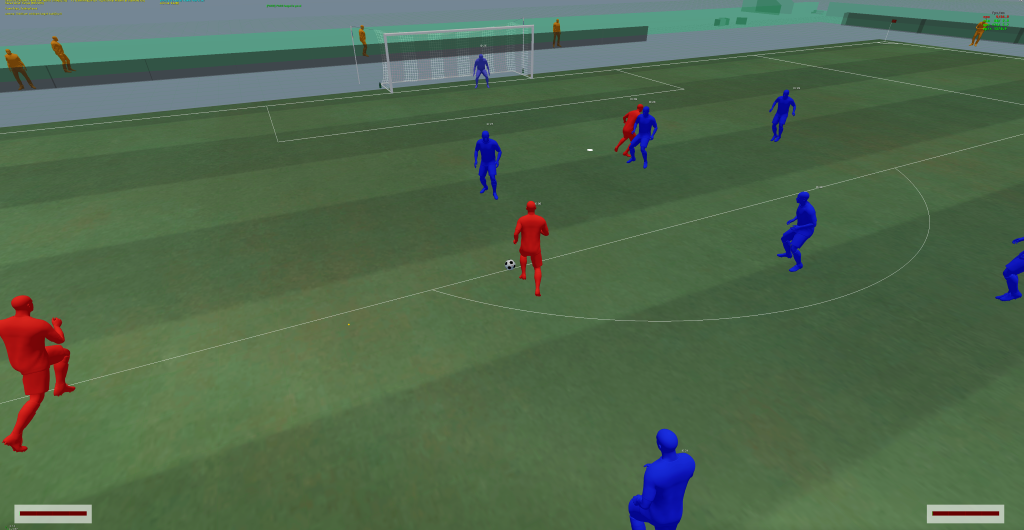}
    \caption{
    \textbf{Example of a test in the automatic expert-authored test suite.} In this test, the striker (in red, in the foreground) is threatening the center position of the goal net, and goalkeeper is very close to the goal line. In this type of situations, the goalkeeper should move forward. For this reason, in this situation the test checks whether the agent is moving towards the red striker.}
    \label{fig:qual_eval_2}
\end{figure}

\begin{figure}
    \centering
     \includegraphics[width=0.90\linewidth]{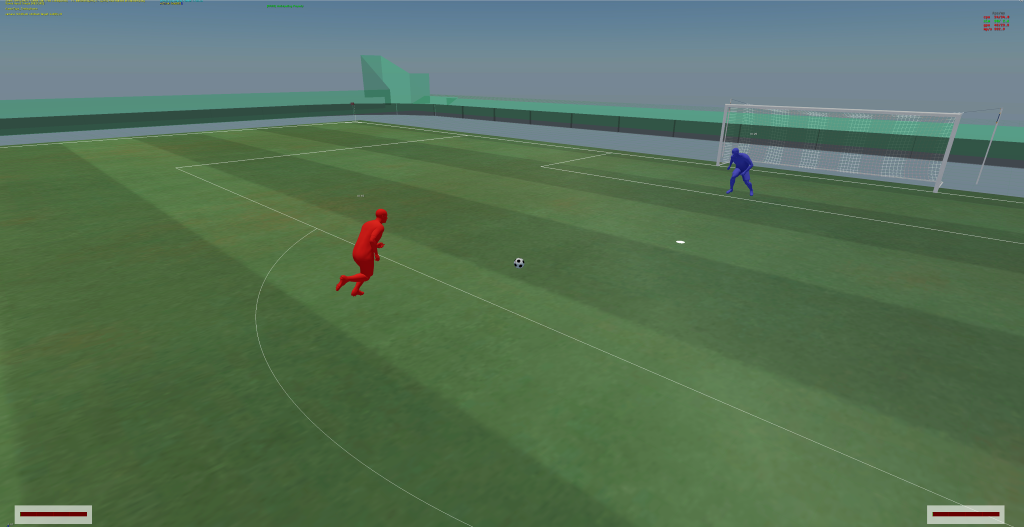}
    \caption{
    \textbf{Example of a test in the automatic expert-authored test suite.} In this test, the striker (in red) and the goalkeeper (in blue) are far from the ball, that is moving towards the goal line. It is possible for the human goalkeeper to catch the ball before the striker, thus this test checks if the agent anticipates the striker.}
    \label{fig:qual_eval_3}
\end{figure}

\begin{figure}
    \centering
     \includegraphics[width=0.90\linewidth]{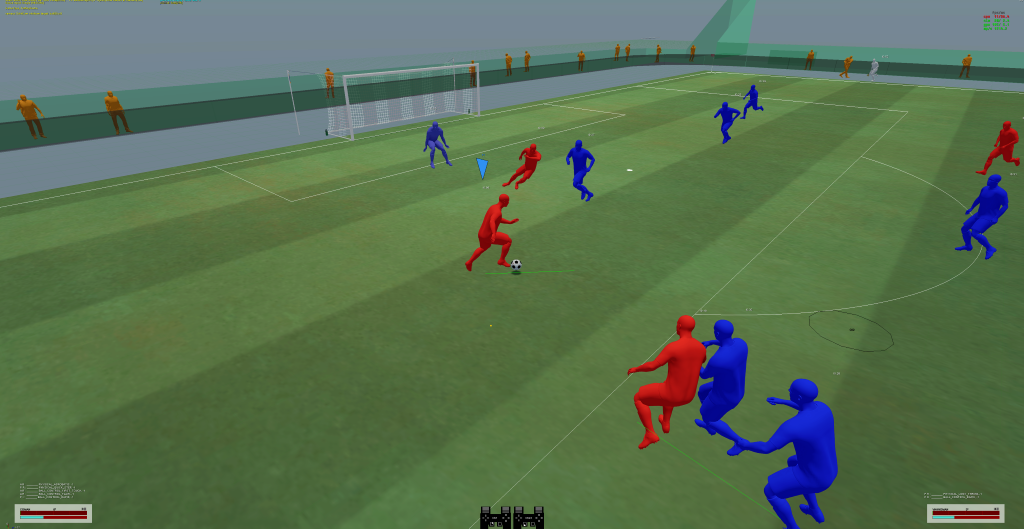}
    \caption{
    \textbf{Example of a test in the expert-authored test suite.} In this test, the goalkeeper is far from the goal net and out of standard position, leaving time and space for the striker to score. Therefore, this test checks if the goalkeeper is going towards the goal net repositioning itself in the standard position.}
    \label{fig:qual_eval_4}
\end{figure}

Figures \ref{fig:qual_eval_1} to \ref{fig:qual_eval_4} show examples of tests and their completion condition used in the expert-authored test suite. We use a total of 344 tests for evaluating the agent, with different completion conditions.

\clearpage

\clearpage

\clearpage

\section{Additional Results in MuJoCo}
\label{app:mujoco}
While motivated by game production constraints, we evaluate generalization to standard continuous control benchmarks. To assess its generality, we test the method on standard MuJoCo continuous control tasks under a 100,000-step budget. These experiments isolate the effect of the training procedure independently of scenario design or domain-specific rewards. We use D4RL \textit{medium} datasets~\citep{d4rl} to simulate sub-optimal samples as our offline datasets and we do not use curriculum learning. The tasks we consider are: \textit{Hopper}, \textit{HalfCheetah}, \textit{Ant}, and \textit{Humanoid}. Table~\ref{tab:mujoco} shows the results. Our approach achieves better results than the standard SAC in all tasks except the Humanoid task. For the latter, we hypothesize that the combination of hard resets and high replay-ratio optimization may be detrimental for stable representation learning in high-dimensional control tasks that require long-horizon coordination. Unlike the other environments, Humanoid exhibits more complex dynamics and higher sensitivity to value estimation noise, which may increase instability introduced by resets. 

\begin{table*}[h]
    \begin{center}
    \scalebox{0.9}{
    \begin{tabular}{lcr}
        \toprule 
        \textbf{Task}   &   \textbf{SAC}                    &   \textbf{Our Method}             \\     
        \midrule 
        Hopper          &   $1830.02 \pm 479.80$            &   $\mathbf{1911.12 \pm 409.40}$   \\
        HalfCheetah     &   $1401.39 \pm 140.15$            &   $\mathbf{4552.16 \pm 129.32}$    \\
        Ant             &   $1253.02 \pm \phantom{0}36.27$  &   $\mathbf{1767.12 \pm 425.99}$   \\
        Humanoid        &   $\mathbf{2059.02 \pm 555.02}$   &   $604.83 \pm \phantom{0}48.03$   \\
        
        \bottomrule
    \end{tabular}
  }
  \caption{\textbf{Additional Results on MuJoCo 100K.} For this experiment, we limit our budget to 100K samples, similar to \citet{td7}. We compare our method with the standard SAC algorithm. Our method achieves better results in all tasks except Humanoid.}
  \label{tab:mujoco} 
  \end{center}
\end{table*}

\end{document}